%% file: main.tex
\documentclass[journal]{IEEEtran}
\usepackage{amsmath,amsfonts}
\usepackage{algorithm}
\usepackage{array}
\usepackage{textcomp}
\usepackage{stfloats}
\usepackage{url}
\usepackage{verbatim}
\usepackage{graphicx}
\usepackage{cite}
\usepackage{xspace}
\usepackage{mathtools}
\newcommand{\ie}{\textit{i.e.}\xspace}
\newcommand{\eg}{\textit{e.g.}\xspace}
\usepackage{pifont}
\usepackage{orcidlink}
\usepackage{multirow}
\usepackage{adjustbox}
\usepackage{booktabs}
\usepackage{color, colortbl}
\usepackage{bbding}
\usepackage{subcaption}
\usepackage{makecell}
\usepackage{algpseudocode}
\renewcommand{\algorithmicrequire}{\textbf{Input:}}
\renewcommand{\algorithmicensure}{\textbf{Output:}}

\begin{document}

\title{Neural Antidote: Class-Wise Prompt Tuning for Purifying Backdoors in CLIP}

\author{
Jiawei Kong\orcidlink{0009-0001-4879-0668},
Hao Fang\orcidlink{0009-0004-0271-6579},
Sihang Guo\orcidlink{0009-0007-7858-9213},
Chenxi Qing\orcidlink{0009-0003-3370-9737}, 
Kuofeng Gao\orcidlink{0000-0002-5667-8238}, 
Bin Chen\orcidlink{0000-0002-4798-230X}, ~\IEEEmembership{Member, IEEE}, \\
% Bin Wang\orcidlink{}, \\
Shu-Tao Xia\orcidlink{0000-0002-8639-982X}, ~\IEEEmembership{Member, IEEE},
Ke Xu\orcidlink{0000-0003-2587-8517}, ~\IEEEmembership{Fellow, IEEE}

\thanks{Jiawei Kong, Hao Fang, Chenxi Qing, Kuofeng Gao, and Shu-Tao Xia are with the Tsinghua Shenzhen International Graduate School, Tsinghua University, Shenzhen, Guangdong 518055, China (e-mail: \{kjw25, fangh25, qcx25, gkf21\}@mails.tsinghua.edu.cn; xiast@sz.tsinghua.edu.cn).}
\thanks{Sihang Guo and Bin Chen and with the School of Computer Science and Technology, Harbin Institute of Technology, Shenzhen, Guangdong 518055, China (e-mail: 220110225@stu.hit.edu.cn; chenbin2021@hit.edu.cn).}
\thanks{Ke Xu is with the Department of Computer Science and Technology, Tsinghua University, Beijing 100084, China (e-mail: xuke@tsinghua.edu.cn).}
\thanks{Jiawei Kong and Hao Fang contributed equally to this paper.}
\thanks{Corresponding author: Bin Chen (e-mail: chenbin2021@hit.edu.cn).}

}



\maketitle

\input{sections/0_abstract}
\input{sections/1_Introduction}
\input{sections/2_Related_Work}
\input{sections/3_Problem_Formulation}
\input{sections/4_Method}

\input{sections/5_Experiments}
\input{sections/6_Conclusion}

\bibliographystyle{IEEEtran}
\bibliography{references}

\input{sections/Appendix}

\end{document}

%% file: sections/0_abstract.tex
\begin{abstract}
While pre-trained Vision-Language Models (VLMs) such as CLIP exhibit impressive representational capabilities for multimodal data, recent studies have revealed their vulnerability to backdoor attacks. To alleviate the threat, existing defense strategies primarily focus on fine-tuning the entire suspicious model. However, the substantial model parameters increase the difficulty of reaching a stable and consistent optimization direction, limiting their resistance against state-of-the-art attacks and often resulting in a degradation of clean accuracy. To address this challenge, we propose Class-wise Backdoor Prompt Tuning (CBPT), an efficient and effective defense mechanism that operates on text prompts to indirectly purify poisoned CLIP. Specifically, we first employ the advanced contrastive learning via carefully crafted positive and negative samples, to effectively invert the backdoor triggers that are potentially adopted by the attacker. Once the dummy trigger is established, we leverage three well-designed loss functions to optimize these class-wise text prompts, modifying the model's decision boundary and further reclassifying the feature regions affected by backdoor triggers. Extensive experiments demonstrate that CBPT significantly mitigates backdoor threats while preserving model utility, e.g. an average Clean Accuracy (CA) of 58.83\% and an Attack Success Rate (ASR) of 0.39\% across seven mainstream backdoor attacks. These results underscore the superiority of our prompt purifying design to strengthen CLIP's robustness against backdoor attacks.
\end{abstract}


%% file: sections/1_Introduction.tex
\section{Introduction}
\label{sec:intro}

\begin{figure}[t]
\begin{center}
\includegraphics[width=\linewidth]{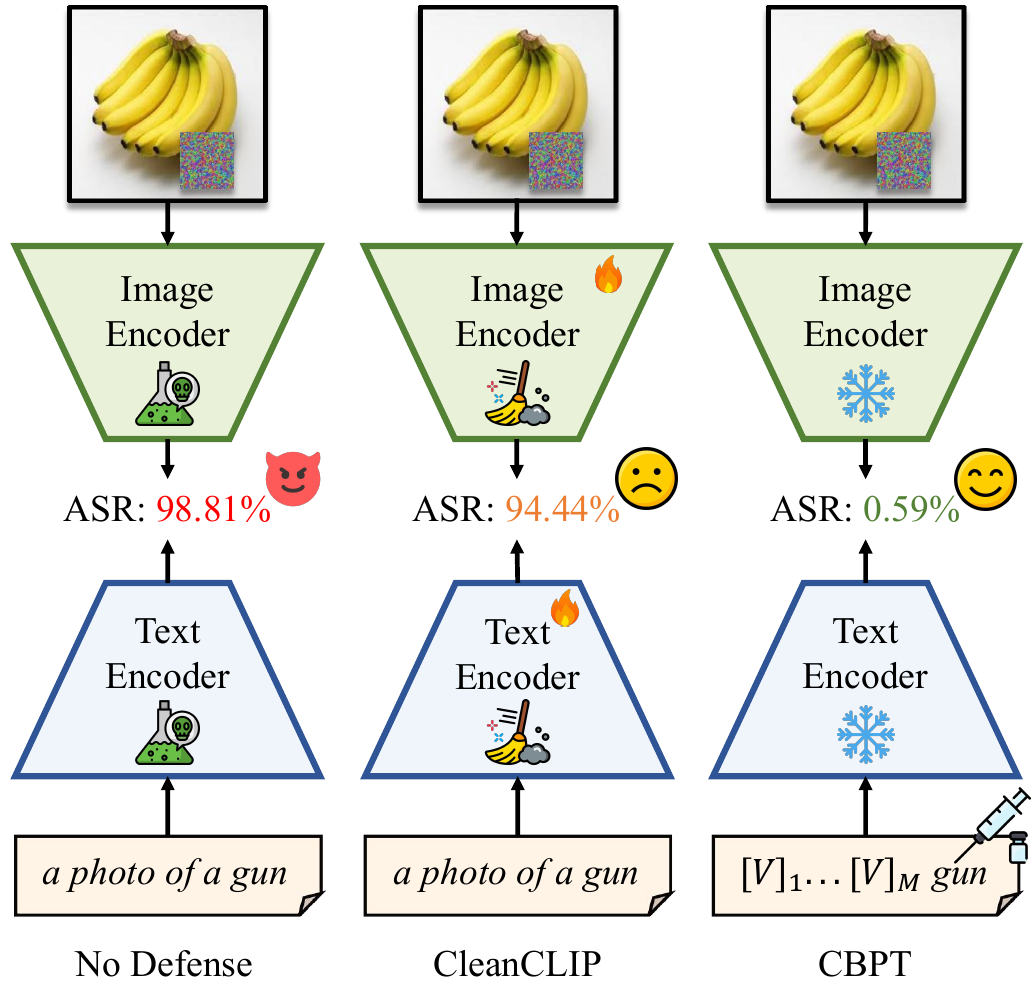}
\end{center}
\caption{Illustration of our proposed CBPT in comparison with previous backdoor defenses. Existing methods such as CleanCLIP \cite{bansal2023cleanclip} finetune the entire backdoored model on a clean dataset, yet typically obtain only a marginal decrease in ASR against the state-of-the-art (SOTA) attacks. In contrast, CBPT freezes the model weights and adopts parameter-efficient prompt tuning to learn robust text representations as neural antidotes, achieving superior defense effectiveness.}
\label{fig:intro}
\end{figure}

Large pre-trained Vision-Language Models (VLMs), such as CLIP \cite{radford2021learning}, ALBEF \cite{li2021align}, TCL \cite{yang2022vision}, have achieved remarkable performance across a wide range of multimodal tasks, \eg, Image-Text Retrieval (ITR) \cite{wang2016comprehensive}, Visual Grounding (VG) \cite{hong2019learning}. Trained on large-scale web-crawled datasets in a self-supervised paradigm, VLMs effectively bridge the modality gap by encoding data into a well-aligned feature space, which in turn provides high-quality representations for various downstream applications. 
Moreover, under the increasingly popular pretrain-and-finetune paradigm, users with limited computational resources can directly download open-source pre-trained VLMs and fine-tune them on smaller, private datasets, thereby striking a trade-off between efficiency and performance.

Despite their remarkable success, VLMs remain highly susceptible to backdoor attacks \cite{liang2024badclip, bai2024badclip}. By injecting specific triggers into a small portion of training data, attackers can poison the training process and implant covert backdoors into the target model. Such poisoned models behave normally on benign inputs, but exhibit attacker-specified behaviors when the trigger is present. 
This vulnerability is particularly alarming in real-world deployments, especially in security-critical domains such as autonomous driving \cite{eykholt2018robust, zhang2022towards}, financial systems \cite{sarkar2018robust}, and facial recognition \cite{xue2021backdoors, le2024comprehensive}, where risks are amplified by their severity and stealthiness.
Even more concerning, the pretrain-and-finetune paradigm may accelerate the propagation of backdoors: once an open-source pre-trained VLM is compromised, its fine-tuned variants are highly likely to inherit the backdoor due to its persistence.

To mitigate backdoor threats in CLIP, several defenses have been proposed to purify maliciously implanted backdoors. Mainstream approaches, such as FT and CleanCLIP \cite{bansal2023cleanclip}, typically fine-tune the entire suspicious model with an auxiliary clean dataset. Unfortunately, these methods provide limited resistance against state-of-the-art attacks, \eg, TrojVQA \cite{walmer2022dual} and BadCLIP \cite{liang2024badclip}. 
% One possible reason for their failure is that the massive parameters in VLMs facilitate the establishment of strong connections between the backdoor triggers and pre-defined outputs, which brings great challenges to unlearning these robust backdoors. Additionally, directly fine-tuning the entire VLM exhibits less effectiveness since the substantial model parameters increase the difficulty of reaching a stable and consistent optimization direction.
% Another disadvantage is that these methods simply utilize benign loss functions to fine-tune the model, totally overlooking the incorporation of information from the attacker-adopted triggers, which further leads to an indirect and less effective purifying direction.
We attribute their failure to the following three key factors: 
\ding{182} The massive parameters in CLIP facilitate the establishment of strong connections between the backdoor triggers and pre-defined outputs, which brings great challenges to unlearning these robust backdoors.
\ding{183} Directly fine-tuning the entire CLIP exhibits less effectiveness since the enormous parameter space increases the difficulty of reaching a stable and consistent optimization direction.
\ding{184} These methods rely solely on benign loss functions to fine-tune the poison model, while completely overlooking the incorporation of information from attacker-adopted triggers, which further leads to an indirect and less effective purifying process.

Building on these insights, we propose Class-wise Backdoor Prompt Tuning (CBPT), an efficient prompt tuning method based on bi-level optimization that learns class-wise, backdoor-robust text prompts for CLIP while keeping the model weights frozen, as illustrated in Fig. \ref{fig:intro}. Motivated by the observation that adversarial noise often exhibits strong similarities to backdoor triggers due to the existence of backdoor shortcuts \cite{wang2020practical, zhu2024neural}, we first optimize adversarial perturbations towards a simulated backdoor class and then exploit the learned adversarial perturbations as dummy triggers with functionality similar to those adopted by attackers.   
% Additionally, we employ the powerful contrastive learning technique to perform effective trigger inversion, \ie, generating perturbations that push adversarial embeddings away from the original ones while pulling them closer to embeddings of carefully crafted positive samples from the target class. Guided by the simulated triggers, we then design three loss functions to optimize the class-wise text prompts using clean samples and poisoned samples injected with the dummy trigger, strengthening CLIP's backdoor robustness while preserving its benign utility. 
Additionally, inspired by the strong alignment capability of contrastive learning, we employ this powerful technique to achieve effective trigger inversion, \ie, generating perturbations that push adversarial embeddings away from their original counterparts while pulling them closer to embeddings of carefully constructed positive samples from the target class. Leveraging these simulated triggers, we further design three tailored loss functions to optimize the class-wise text prompts with both clean samples and poisoned samples injected with the dummy trigger, thereby strengthening CLIP's backdoor robustness while maintaining its benign utility. 

We conduct extensive experiments to benchmark CBPT against seven backdoor attacks and two defenses. The significant reduction in attack success rates (ASR) coupled with the preservation of clean accuracy (CA), demonstrates the superiority of our proposed defense over prior defenses. To further assess the generalizability of our approach, we evaluate CBPT under cross-domain scenarios where the fine-tuning dataset exhibits domain shifts relative to the test data.
In addition, we investigate performance across varying data and parameter scales, including 1-, 2-, 4-, 8-, and 16-shot learning as well as 1-, 2-, 4-, 8-, and 16-token settings, highlighting the data- and parameter-efficiency of our approach.
% We further investigate three context positions to assess their impact on defense effectiveness.

In summary, our main \textbf{contributions} are as follows:
\begin{itemize}
    \item We propose CBPT, a novel class-wise prompt tuning method that inverts potential backdoor triggers and optimizes class-wise prompts to boost backdoor robustness.
    \item We formulate a contrastive paradigm for effective and efficient trigger inversion, based on which we design three novel loss functions for robust text prompt tuning.
    \item Extensive experiments demonstrate that our method effectively purifies SOTA backdoor attacks, including both unimodal and multimodal attacks on CLIP models, outperforming SOTA defenses by a substantial margin.
\end{itemize}

%% file: sections/2_Related_Work.tex
\section{Related Work}
\label{sec:related work}

\subsection{Backdoor Attacks and Defences}
Among various security problems \cite{fang2023gifd, fang2024clip, fang2024one, fang2024privacy, fang2025retrievals, fang2025grounding, fang2025your, kong2025wolf, yu2025gi}, backdoor stands out as a particularly important type.

\textbf{Backdoor attack} is an in-training threat in which the adversary poisons a small subset of training samples, thereby implanting a concealed Trojan into the model. While the model behaves normally on benign inputs, it produces attacker-specified outputs once the backdoor is activated by a specific trigger. Since the pioneer work \cite{gu2019badnets} revealed the backdoor vulnerability of CNNs, extensive research \cite{chen2017targeted, barni2019new, li2021invisible, nguyen2021wanet} has sought to improve attack success rate and improve trigger stealth. For instance, Blended \cite{chen2017targeted} integrates triggers into images via weighted addition, while SIG \cite{barni2019new} perturbs target-class samples without label poisoning, significantly improving imperceptibility. In the domain of VLMs, TrojVQA \cite{walmer2022dual} embeds triggers into both modalities, activating the attack only when both triggers are presented. BadCLIP \cite{liang2024badclip} leverages dual-embedding-guided trigger optimization to align the poisoned samples with target features, achieving state-of-the-art attack performance. The growing success of such attacks underscores the urgent need for a robust and comprehensive defense.

\textbf{Backdoor defense} aims to mitigate potential backdoor threats and is typically categorized into in-training and post-processing approaches based on their operational stage. In-training defenses, which have access to the training dataset, identify suspicious samples through methods such as transformation sensitivity \cite{chen2022effective} and feature clustering \cite{huang2022backdoor}. Post-processing defenses, in contrast, primarily focus on techniques such as trigger reversion \cite{wang2019neural, chen2019deepinspect}, model pruning \cite{liu2018fine, wu2021adversarial, li2023reconstructive} and fine-tuning \cite{li2021neural, zeng2021adversarial}. For VLMs, CleanCLIP \cite{bansal2023cleanclip} fine-tunes the model on a clean dataset with a combination of multimodal contrastive loss and unimodal self-supervised loss, while RoCLIP \cite{yang2024robust} constructs a large caption pool to pair images with their most semantically similar captions. 
More recently, SafeCLIP \cite{yang2024better} separates data into safe and risky sets after unimodal contrastive learning warm-up and applies different loss functions to the two sets.
However, current methods exhibit limited resistance to advanced attacks and often degrade performance on benign data. 
To address these limitations, we propose leveraging prompt tuning, a highly effective technique in VLMs, to efficiently eliminate backdoors while preserving model utility to a significant extent.

\subsection{Prompt Tuning}
Prompt tuning originated in Natural Language Processing (NLP), where prompts act as model-friendly input formats. Rather than altering internal model weights, prompt tuning adapts pre-trained models to downstream tasks by optimizing prompts, thereby substantially reducing computational costs. CoOp \cite{zhou2022learning} pioneers the application of prompt tuning in VLMs, particularly for adapting CLIP to image recognition. Building on this, CoCoOp \cite{zhou2022conditional} introduces input-conditional prompts, enhancing generalization to unseen classes via a lightweight, learnable neural network. Beyond image recognition, prompt tuning has also been extended to diverse domains, including text-to-image synthesis \cite{tao2023galip}, semantic segmentation \cite{zhou2023zegclip} and adversarial robustness \cite{li2024one}. Despite these advances, its potential in backdoor purification remains largely unexplored. In this paper, we address this gap by investigating how prompt tuning can enhance robustness against backdoor attacks.

%% file: sections/3_Problem_Formulation.tex
\section{Preliminaries}
\label{preliminariy}

\subsection{Review of CLIP}
CLIP-like VLMs aim to project multimodal inputs into a shared, highly aligned feature space. Typically, such models consist of an image encoder $f_v(\cdot)$ and a text encoder $f_t(\cdot)$, parameterized by $\boldsymbol{\theta}_v$ and $\boldsymbol{\theta}_t$ respectively. For an input image-text pair $(v, t)$, the embeddings are obtained as::
\begin{eqnarray}
    e_v=f_v(v;\boldsymbol{\theta}_v),\ \ e_t=f_t(t;\boldsymbol{\theta}_t).
\end{eqnarray}
The correlation between the image and text can be quantified by the cosine similarity of their respective embeddings:
\begin{eqnarray}
    s(v,t)=\cos(e_v, e_t).
\end{eqnarray}

During training, CLIP bridges different modalities through contrastive learning. Given a batch of image-text pairs $\{v_i, t_i\}_{i=1}^{N}$, the objective is to maximize similarity for matched pairs while minimizing it for mismatched ones:
% CLIP maximizes the similarity between matching pairs while minimizing that for unrelated pairs:
\begin{eqnarray}
\begin{split}
\mathop{\arg\min}\limits_{\boldsymbol{\theta}_v, \boldsymbol{\theta}_t} &-\frac{1}{2N}\sum_{i=1}^{N} \log \left( \frac{\exp(s(v_i, t_i)/\tau)}{\sum_{j=1}^{N} \exp(s(v_i, t_j)/\tau)} \right) \\
&-\frac{1}{2N}\sum_{i=1}^{N} \log \left( \frac{\exp(s(v_i, t_i)/\tau)}{\sum_{j=1}^{N} \exp(s(v_j, t_i)/\tau)} \right). 
\end{split}
\end{eqnarray}
Benefiting from pre-training on large-scale web-crawled datasets, CLIP exhibits strong generalization ability and can be readily adapted to various downstream tasks. For example, in zero-shot image classification, the label is predicted as the one with the highest similarity between the image embedding and the embeddings of candidate label descriptions:
\begin{eqnarray}
\mathop{\arg\max}\limits_{j} s(v, t_j).
\end{eqnarray}

In practical applications, handcrafted templates such as ''\texttt{a photo of a [CLASS]}'',  are often used to improve adaptability and stability \cite{huang2023sentence, radford2021learning}. Here, the phrase ''\texttt{a photo of a}'' provides contextual cues that guide decision-making. However, handcrafted prompts typically rely on expert intuition and are restricted to discrete words. To address these limitations, prompt tuning has emerged as a more flexible alternative by learning prompt vectors in a continuous space, achieving substantial performance gains \cite{zhou2022learning}. Formally, assuming that the context contains $p$ preceding tokens and $M-p$ following tokens, the encoded text feature can be expressed as:
\begin{eqnarray}
    e_t = f_t(\operatorname{concat}([V_{\text{front}}, \operatorname{tokenize}([\operatorname{CLASS}]), V_{\text{end}}])),
    \label{eq:context}
\end{eqnarray}
where $V_{\text{front}} \in \mathbb{R}^{p \times d}$, $V_{\text{end}} \in \mathbb{R}^{(M-p) \times d}$ are learnable parameters, and $d$ denotes the dimension of the token vector.

\subsection{Threat Model}
Since existing studies on both attacks and defenses predominantly focus on image classification, we follow the same paradigm and restrict our scope to backdoor purification in this setting. To implant a backdoor into VLMs, the attacker first specifies a target class $T$, and selects a subset $\mathcal{D}_{sub}=\{v_i, t_i\}_{i=1}^{m}$ from the training set $\mathcal{D}_{train}$, which is subsequently embedded with attacker-specified triggers:
\begin{eqnarray}
    \hat{v}_i=v_i \oplus \Delta_v, \ \ \hat{t}_i=t_i \oplus \Delta_t,
\end{eqnarray}
where $\Delta_v$ and $\Delta_t$ denote the visual and text triggers respectively, and $\oplus$ represents the fusion operation. The poisoned samples $\mathcal{D}_{poi}=\{\hat{v}_i, \hat{t}_i\}_{i=1}^{m}$ are combined with the remaining benign samples to construct the modified training set: $\mathcal{D}_{train}'=\mathcal{D}_{poi}\cup(\mathcal{D}_{train}-\mathcal{D}_{sub})$. Trained on this malicious dataset, the model behaves normally on benign inputs, but exhibits harmful behavior once the triggers appear:
\begin{eqnarray}
    \mathop{\arg\max}\limits_{j} s(v_i, t_j)=i, \ \ \mathop{\arg\max}\limits_{j} s(\hat{v}_i, \hat{t}_j)=T.
\end{eqnarray}

%% file: sections/4_Method.tex
\section{Method}
\label{method}

\begin{figure}[t]
\begin{center}
\includegraphics[width=0.9\linewidth]{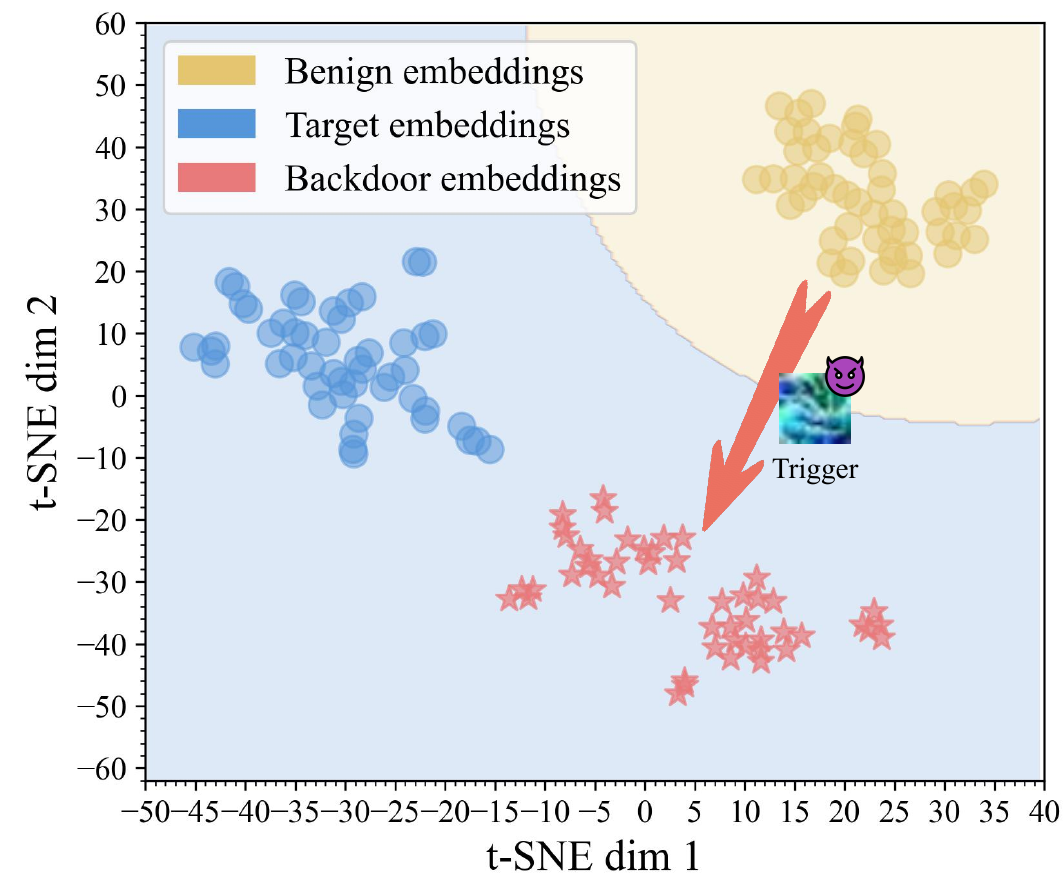}
\end{center}
\caption{The visualization of visual embeddings. Blue dots denote samples from the backdoor target class, yellow dots represent benign samples, and red stars indicate samples embedded with malicious triggers.}
\label{fig:motivation}
\end{figure}

In this section, we first present the motivation for defense by analyzing existing backdoor attacks. We then introduce Class-wise Backdoor Prompt Tuning (CBPT), a bi-level optimization framework that incorporates inner optimization for trigger inversion and outer optimization for prompt tuning, both designed to effectively purify backdoors.

\begin{figure*}
\centering
\includegraphics[width=0.9\linewidth]{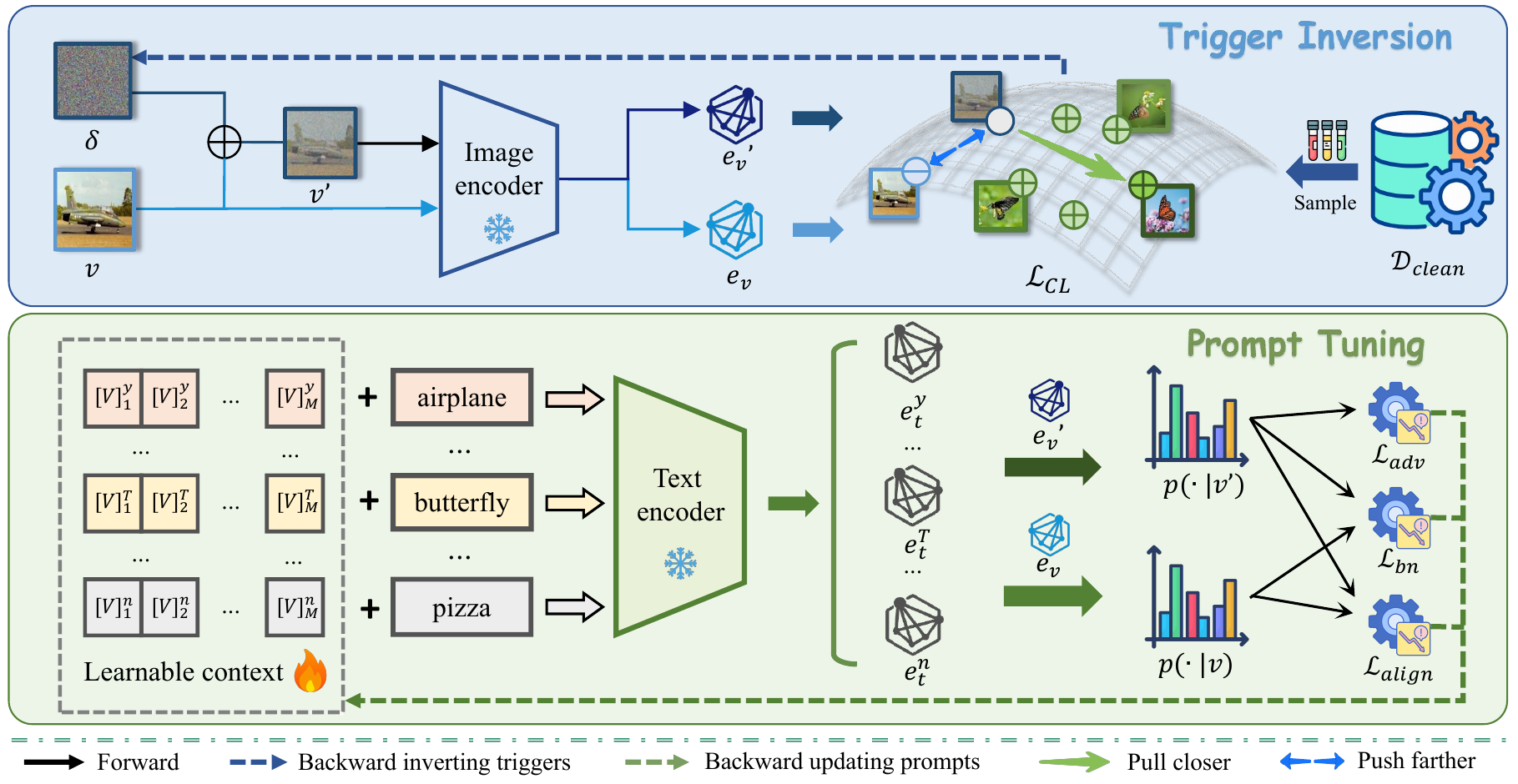}
\caption{Illustration of the proposed Class-wise Backdoor Prompt Tuning (CBPT) method. 
In the inner optimization stage of trigger inversion, instance-specific noises are learned to simulate backdoor triggers via contrastive learning. In the outer optimization stage of prompt tuning, three novel loss functions are introduced to tune the context.
Throughout both stages, both the image and text encoders remain frozen, while the learnable context is class-specific.} 
\label{pipeline.pdf}
\end{figure*}

\subsection{Motivation}
Existing backdoor attacks, whether unimodal or multimodal, primarily target the visual modality \cite{gu2019badnets, liang2024badclip}. In other words, their success largely depends on the precise manipulation of the image features. Using BadCLIP \cite{liang2024badclip} as a case study, we employ t-SNE \cite{van2008visualizing} to visualize three types of samples: benign images from the backdoor target class, benign images from a randomly selected class, and the corresponding images embedded with malicious triggers, as illustrated in Fig. \ref{fig:motivation}. From this visualization, we derive two key observations:

\ding{182} The success of backdoor activation stems from the dramatic shift of benign image features towards the target class region, induced by implanting an attacker-specified trigger.

\ding{183} Although the malicious image features fall within the decision boundary of the target class, they still exhibit a notable deviation from the genuine samples of that class.

 % In particular, a carefully tuned text representation can shift the decision boundary in a way that suppresses the influence of triggers without compromising recognition of clean samples. 
 % This property ensures that purification not only mitigates the backdoor threat but also maintains the model’s ability to generalize well on benign data.
 % Importantly, this separability indicates that backdoor purification can be achieved without substantially altering the underlying feature distribution, ensuring robustness and stability during deployment.

The above findings provide a solid foundation for backdoor purification via prompt tuning. Since CLIP performs classification by measuring the similarity between image features and text features of candidate classes, text features play a pivotal role in shaping the decision boundary. This motivates us to construct a more precise boundary that effectively separates malicious samples from benign target-class images. 
Moreover, the observation that intra-cluster distances among benign features are significantly smaller than the distances between benign and malicious feature clusters highlights the great separability between malicious and benign samples.
Importantly, this inherent separability indicates that backdoor purification can be achieved without substantially altering the underlying feature distribution, thereby preserving the overall model utility and ensuring robustness during deployment.

\subsection{Class-wise Backdoor Prompt Tuning}
Motivated by the above observations, we propose CBPT, which leverages a clean labeled dataset $\mathcal{D}_{clean}=\{v_i, y_i\}_{i=1}^S$ to learn backdoor-robust prompts for a suspicious CLIP model. As depicted in Fig. \ref{pipeline.pdf}, CBPT comprises two key optimization stages, \ie, trigger inversion and prompt tuning. In the trigger inversion stage, we elaborately construct positive and negative samples and adopt a contrastive learning paradigm to invert the trigger. In the subsequent prompt tuning stage, the simulated trigger features are employed to optimize prompt vectors, thereby refining the decision boundary and enhancing resilience against unseen malicious triggers. The pseudocode of CBPT is provided in Algorithm \ref{alg:CBPT}. 
We then specify the parameterization of class-wise prompts and elaborate on the design of the inner and outer optimization processes. 

\textbf{Prompt parameterization.}
To improve clarity, we first reformulate the learnable context in Eq. (\ref{eq:context}) as a parameterized structure. We adopt class-wise prompts to enhance the purification effect against backdoor attacks, assigning each class a distinct set of context vectors. Specifically, the context of a given class $[\text{CLASS}]_j$ consists of $M$ tokens, each represented by $[V]_i^j \in \mathbb{R}^d$ for $i \in \{1,\ldots, M\}$. These context vectors are then concatenated with the tokenized class vector $[C_j]$ to construct the “text” $t_j$, which is then passed through the text encoder $f_t(\cdot)$ to obtain the text embedding $e_t^j$:
\begin{eqnarray}
    e_t^j=f_t([V]_1^j\ldots[V]_p^j[C]_j[V]_{p+1}^j\ldots[V]_M^j),
\end{eqnarray}
where $p$ denotes the number of prefix tokens. 
% Following prior works \cite{zhou2022learning, zhou2022conditional, li2024one}, we evaluate our method under three position settings for $p$: \textit{front}, \textit{middle} and \textit{end} (\ie, $p=0,M/2,M$), with the \textbf{end} position serving as the default in our main experiments.

% Formally, this requires ensuring that

\renewcommand{\algorithmicrequire}{\textbf{Input:}}  % Use Input in the format of Algorithm
\renewcommand{\algorithmicensure}{\textbf{Output:}} % Use Output in the format of Algorithm

\begin{algorithm}[h]
  \caption{Pseudocode of CBPT} \label{CBPT}
  \begin{algorithmic}[1]
    \Require
      $\mathcal{D}_{clean}$: clean fine-tuning dataset;
      $f_v(\cdot), f_t(\cdot)$: image encoder and text encoder of the suspicious CLIP;
      $\mathcal{T}_0$: number of warm-up epochs;
      $\mathcal{T}$: number of training epochs;
      $\mathcal{T}_{inner}$: number of inner loops.
    \Ensure
       Class-wise prompts $[V] \in \mathbb{R}^{n \times M \times d}$.
       \State Initialize the prompt $[V]$ with Gaussian distribution;
       \State Warm-up: Train $[V]$ with $\mathcal{D}_{clean}$ for $\mathcal{T}_0$ epochs;
       \For{$i \leftarrow 0$ to $\mathcal{T}-1$}
            \State \parbox[t]{\dimexpr\linewidth-\algorithmicindent}{\texttt{//Outer loop}}
            \For{$(v,y) \in \mathcal{D}_{clean}$}
                \State Initialize noise $\delta$ to zero;
                \State Approximate backdoor class $y'$ by Eq. (\ref{eq:backdoor_class});
                % \State Obtain positive samples $\{v_{pos,i}\}_{i=1}^{b}$ and negative samples $\{v_{neg,i}\}_{i=1}^{b}$;
                \State Obtain positive samples $v_{pos}$ from $\mathcal{D}_{clean}$;
                \State Obtain negative samples $v_{neg}\leftarrow v$;
                \For{$j \leftarrow 0$ to $\mathcal{T}_{inner}-1$}
                    \State \parbox[t]{\dimexpr\linewidth-\algorithmicindent}{\texttt{//Inner loop}}
                    \State Compute $\mathcal{L}_{CL}$ with $(v, \delta)$ by Eq. (\ref{eq:loss_CL});
                    \State Perform backward pass and update $\delta$;
                \EndFor
                \State Compute $\mathcal{L}_{adv}$ with $(v,\delta,y')$ by Eq. (\ref{eq:loss_adv});
                \State Compute $\mathcal{L}_{align}$ with $(v,\delta,y)$ by Eq. (\ref{eq:loss_align});
                \State Compute $\mathcal{L}_{bn}$ with $(v,\delta,y)$ by Eq. (\ref{eq:loss_bn});
                \State Perform backward pass and update $[V]$;
             \EndFor   
       \EndFor
  \State Return: Class-wise prompts $[V]$;
  \end{algorithmic}
  \label{alg:CBPT}
\end{algorithm}

In essence, our main objective is to optimize the context vectors $[V]_i^j$ (for $i \in \{1,\ldots, M\}$ and $j \in \{1, \ldots, n\}$, where $n$ indicates the number of classes) so as to yield more robust text features resistant to backdoor attacks.
Formally, this objective can be explicitly expressed as ensuring that
% $\mathop{\arg\max}\limits_{j} s(\hat{v}_i, \hat{t}_j)\neq T$.
\begin{eqnarray}
\mathop{\arg\max}\limits_{j} s(\hat{v}_i, \hat{t}_j)\neq T,
\end{eqnarray}
where $T$ denotes the malicious target class.

\textbf{Trigger inversion.}
Due to backdoor shortcuts, per-image adversarial perturbations closely resemble images with malicious triggers at the feature level \cite{wang2020practical}. Leveraging this property, we optimize instance-specific noise, with the same dimensionality as the input image, to invert potential triggers and simulate the malicious feature region.
Formally, given a clean image with label $y$ from $\mathcal{D}_{clean}$, we generate a perturbed image $v+\delta$ and encode both using the image encoder $f_v(\cdot)$:
\begin{eqnarray}
    e_v=f_v(v), \ \ e_v'=f_v(v+\delta).
\end{eqnarray}

Meanwhile, each class vector concatenated with its corresponding prompt is encoded by the text encoder $f_t(\cdot)$, producing a set of text embeddings $\{e_t^i\}_{i=1}^n$. With these components, we adopt an instance-specific and dynamic strategy to approximate the backdoor target class $T$ as:
\begin{eqnarray}
    T \approx y'=\mathop{\arg\max}\limits_{k\neq y} {s(e_v, e_t^k)},
    \label{eq:backdoor_class}
\end{eqnarray}
which has been validated in prior work \cite{zhu2024neural}. Subsequently, our goal is to optimize $\delta$ so that the perturbed image $v+\delta$ is drawn closer to samples of the target class, ultimately falling within its decision boundary. 
Different from previous PGD-based methods \cite{madry2017towards} that rely on reliable supervision signals, the textual features in our prompt learning fail to solidly provide precise guidance since the introduced prompt vectors $[V]_i^j$ are randomly initialized and introduce noisy interference in the early optimization, leading to inefficient convergence and suboptimal results. To tackle this, we turn to incorporating clean visual supervision to guide the dummy trigger modeling.

Specifically, we employ the advanced contrastive learning mechanism for more effective and reliable trigger inversion. To formulate the contrastive paradigm, we first select images labeled as $y'$ from the dataset, encode them, and choose the one with the largest feature distance as the positive sample $v_{pos}$, while the original clean image itself serves as the negative sample, \ie, $v_{neg}=v$. Using these crafted samples, we optimize $\delta$ by minimizing $\mathcal{L}_{CL}$ in a contrastive manner, pushing the embedding of the perturbed image apart from the clean one while pulling it towards the target embedding:
\begin{eqnarray}
\begin{split}
    \mathcal{L}_{CL}(v, \delta)&= s(f_v(v+\delta), f_v(v_{neg})) \\ & - s(f_v(v+\delta), f_v(v_{pos})).
    \label{eq:loss_CL}
\end{split}
\end{eqnarray}
% \begin{eqnarray}
% \begin{split}
%     \mathcal{L}_{CL}(v, \delta)= s(f_v(v+\delta), f_v(v_{neg}))  - s(f_v(v+\delta), f_v(v_{pos})).
%     \label{eq:loss_CL}
% \end{split}
% \end{eqnarray}
% \begin{equation}
%     \mathcal{L}_{CL}(v, \delta)=  s(f_v(v+\delta), f_v(v_{neg}))- s(f_v(v+\delta), f_v(v_{pos})).
%     \label{eq:loss_CL}
% \end{equation}
% where $\alpha$ is a hyper-parameter to controlling the balance between the effect of positive and negative samples.

\begin{table*}[htbp]
  \centering
  \caption{The number of fine-tuned parameters (M), Clean Accuracy (\%) and Attack Success Rate (\%) of our CBPT compared with SOTA defenses against seven backdoor attacks on the ImageNet validation set.}
  % \hspace*{-1cm}
  % \small
  % \setlength{\tabcolsep}{3pt} % 调整列间距
  % \resizebox{\textwidth}{!}
  \resizebox{\textwidth}{!}
  {\begin{tabular}{cccccccccccccccc} 
    \toprule
    % \multicolumn{1}{c}{Method}
    \multicolumn{1}{c}{\multirow{2}[0]{*}{Method}}
    & \multicolumn{1}{c}{\multirow{2}[0]{*}{Params}}
    & \multicolumn{2}{c}{BadNet}          
    & \multicolumn{2}{c}{Blended}          
    & \multicolumn{2}{c}{SIG}          
    & \multicolumn{2}{c}{SSBA}          
    & \multicolumn{2}{c}{WaNet}          
    & \multicolumn{2}{c}{TrojVQA}          
    & \multicolumn{2}{c}{BadCLIP} \\ 

    % \cmidrule(lr){2-3} 
    % \cmidrule(lr){4-5} 
    % \cmidrule(lr){6-7} 
    % \cmidrule(lr){8-9} 
    % \cmidrule(lr){10-11} 
    % \cmidrule(lr){12-13} 
    % \cmidrule(lr){14-15}
    \cmidrule(lr){3-4} 
    \cmidrule(lr){5-6} 
    \cmidrule(lr){7-8} 
    \cmidrule(lr){9-10} 
    \cmidrule(lr){11-12} 
    \cmidrule(lr){13-14} 
    \cmidrule(lr){15-16}

    &
    & CA & ASR 
    & CA & ASR 
    & CA & ASR 
    & CA & ASR 
    & CA & ASR 
    & CA & ASR 
    & CA & ASR \\ 
    
    \midrule
    
    No Defense   &  -
    & 58.83    & 96.51    & 59.06    & 97.61    & \textbf{59.3} & 77.73    & 58.33    & 41.66    & 59.15    & 86       & \textbf{58.68} & 97.86    & 58.72    & 98.81 \\
    
    FT        &  102.01
    & 58.62    & 43.52    & 58.24    & 19.72    & 59.08    & 35.13    & 58.24    & 1.72     & 57.76    & 48.5     & 58.09    & 83.08    & 58.45    & 95.68 \\
    
    CleanCLIP   &  102.01
    & 57.83    & 19.4     & 57.78    & 8.11     & 58.65    & 18.35    & 58.12    & 0.65     & 58.71    & 26.47    & 57.88    & 45.78    & 57.71    & 94.44 \\
    
    \cellcolor[gray]{0.9}CBPT   &  \cellcolor[gray]{0.9}\textbf{2.05}  
    & \cellcolor[gray]{0.9}\textbf{58.88} & \cellcolor[gray]{0.9}\textbf{0.08}  
    & \cellcolor[gray]{0.9}\textbf{59.13} & \cellcolor[gray]{0.9}\textbf{0.32} 
    & \cellcolor[gray]{0.9}59.01 & \cellcolor[gray]{0.9}\textbf{0.97} 
    & \cellcolor[gray]{0.9}\textbf{58.51} & \cellcolor[gray]{0.9}\textbf{0.17} 
    & \cellcolor[gray]{0.9}\textbf{59.17} & \cellcolor[gray]{0.9}\textbf{0.35} 
    & \cellcolor[gray]{0.9}58.12 & \cellcolor[gray]{0.9}\textbf{0.27}  
    & \cellcolor[gray]{0.9}\textbf{58.98} & \cellcolor[gray]{0.9}\textbf{0.59} \\ 
    
    \bottomrule
  \end{tabular}
  }
  \label{main_tab}%
\end{table*}

\textbf{Prompt tuning.}
Leveraging the simulated trigger optimized in the inner loop, we then tune the class-wise prompts in the outer loop to enhance their resistance against the trigger feature. For clarity, we denote $p(y|v)$ as the confidence that the CLIP model assigns to image $v$ for label $y$:
% \begin{equation}
%     f(v;y)=\frac{\exp{(s(v, t_y))}} {\sum_{i=1}^{n} {\exp{(s(v, t_i))}}}.
% \end{equation}
\begin{eqnarray}
    p(y|v)=\frac{\exp{(s(v, t_y))}} {\sum_{i=1}^{n} {\exp{(s(v, t_i))}}}.
\end{eqnarray}
To effectively guide the optimization, we design three loss functions: $\mathcal{L}_{adv}$, $\mathcal{L}_{align}$, and $\mathcal{L}_{bn}$, defined as follows: 

\ding{182} \textbf{Adversarial loss $\mathcal{L}_{adv}$.}  
This loss is defined as the negative log-likelihood of the perturbed image for $y'$, guiding the model to push malicious images away from backdoor class:
% \begin{equation}
%     \mathcal{L}_{adv}(v, \delta, y, y')=-\log(1-f(v+\delta; y')).
%     \label{eq:loss_asr}
% \end{equation}
\begin{eqnarray}
    \mathcal{L}_{adv}(v, \delta, y')=-\log(1-p(y'|v+\delta)).
    \label{eq:loss_adv}
\end{eqnarray}

% \ding{183} To reclassify the perturbed image to its ground truth label $y$, we leverage boosted cross entropy to define $\mathcal{L}_{bce}$. The first term is designed to increase the confidence in label $y$ while the second term prevents an excessive increase in confidence for other unrelated classes:
% % \begin{equation}
% % \begin{split}
% %     \mathcal{L}_{bce}(v, \delta, y)=
% %     &-\log(f(v+\delta; y)) \\
% %     &-\log(1-\mathop{\max}\limits_{k\neq y} {f(v+\delta;k)})
% % \end{split}
% % \end{equation}
% \begin{equation}
%     \mathcal{L}_{bce}(v, \delta, y)=
%     -\log(f(v+\delta; y))-\log(1-\mathop{\max}\limits_{k\neq y} {f(v+\delta;k)}).
%     \label{eq:loss_bce}
% \end{equation}

\ding{183} \textbf{Alignment loss $\mathcal{L}_{align}$.} This loss aligns the predicted probability distribution of the perturbed image with that of the benign counterpart using $l_2$ distance. To mitigate the potential deviation in CLIP’s predictions on benign images, we weight the $l_2$ distance by the model’s confidence in the true label of the anchor benign sample. This adaptive weighting mechanism ensures that each sample's contribution is proportional to the reliability of its distribution. 
Formally, $\mathcal{L}_{align}$ is defined as:
% \begin{equation}
%     \mathcal{L}_{align}(v, \delta, y)=
%     f(v;y)\cdot\frac{1}{n}\sum_{i=1}^{n}(f(v;i)-f(v+\delta;i))^2.
%     \label{eq:loss_align}
% \end{equation}
% \begin{equation}
%     \mathcal{L}_{align}(v, \delta, y)=
%     f(v;y)\cdot \mathcal{L}_{2}(f(v|\cdot), f(v+\delta|\cdot)).
%     \label{eq:loss_align}
% \end{equation}
\begin{eqnarray}
    \mathcal{L}_{align}(v, \delta, y)=
    p(y|v)\cdot \mathcal{L}_{2}\Big(p(\cdot|v), p(\cdot|v+\delta)\Big).
    \label{eq:loss_align}
\end{eqnarray}
% \begin{equation}
% \begin{split}
%     \mathcal{L}_{bce}(v, \delta, y)=
%     &-\log(f(v+\delta; y)) \\
%     &-\log(1-\mathop{\max}\limits_{k\neq y} {f(v+\delta;k)})
% \end{split}
% \end{equation}

\ding{184} \textbf{Benign loss $\mathcal{L}_{bn}$.}
To preserve model utility and prevent excessive alterations to the decision boundary, we introduce the benign loss $\mathcal{L}_{bn}$ based on widely-used cross-entropy loss. This loss encourages the model to maintain accurate predictions for both perturbed and non-perturbed images.
% \begin{equation}
%     \mathcal{L}_{bn}(v,\delta, y)=-\log(f(v;y))-\alpha\log(f(v+\delta;y)).
%     \label{eq:loss_bn}
% \end{equation}
\begin{eqnarray}
    \mathcal{L}_{bn}(v,\delta, y)=-\log(p(y|v))-\alpha\log(p(y|v+\delta)).
    \label{eq:loss_bn}
\end{eqnarray}

By jointly optimizing these three objectives, we refine the class-wise prompts so that the decision boundary is precisely adjusted to reassign trigger-induced feature regions while preserving the structural integrity of benign feature regions.

\textbf{Optimization objective.}
% In summary, we formulate our CBPT as a bi-level optimization problem for learning backdoor-robust text prompts:
In summary, we formulate the proposed CBPT as a bi-level optimization problem:
% \begin{eqnarray}
% \begin{split}
%     \mathop{\min}\limits_{[V]} \ &\lambda_1 \mathcal{L}_{adv}(v,\delta^*,y)+\lambda_2\mathcal{L}_{align}(v, \delta^*,y')+\lambda_3\mathcal{L}_{bn}(v, \delta^*,y), \\
%     & \text{s.t.} \ \delta^*=\mathop{\min}\limits_{||\delta||_p \leq \rho} \mathcal{L}_{CL}(v, \delta),\ p \in \{0,1,2,\infty\},
%     \label{eq:general}
% \end{split}
% \end{eqnarray}
% \begin{eqnarray}
% \begin{split}
%     \mathop{\min}\limits_{[V]} \lambda_1 \mathcal{L}_{adv}(v,\delta^*,y' &)+\lambda_2\mathcal{L}_{align}(v, \delta^*,y)+\lambda_3\mathcal{L}_{bn}(v, \delta^*,y), \\
%     & \text{s.t.} \ \delta^*=\mathop{\min}\limits_{||\delta||_2 \leq \rho} \mathcal{L}_{CL}(v, \delta),
%     \label{eq:general}
% \end{split}
% \end{eqnarray}
\begin{eqnarray}
\begin{split}
    \mathop{\min}\limits_{[V]} \lambda_1 \mathcal{L}_{adv}&(v,\delta^* )+\lambda_2\mathcal{L}_{align}( v,\delta^*)+\lambda_3\mathcal{L}_{bn}( v,\delta^*), \\
    & \text{s.t.} \ \delta^*=\mathop{\min}\limits_{||\delta||_2 \leq \rho} \mathcal{L}_{CL}(v, \delta),
    \label{eq:general}
\end{split}
\end{eqnarray}
% \begin{eqnarray}
% \begin{split}
%     \mathop{\min}\limits_{[V]}& \lambda_1 \mathcal{L}_{adv}(v,\delta^*,y' )+\lambda_2\mathcal{L}_{align}(v, \delta^*,y)\\
%     &+\lambda_3\mathcal{L}_{bn}(v, \delta^*,y), \ \  \text{s.t.} \ \delta^*=\mathop{\min}\limits_{||\delta||_2 \leq \rho} \mathcal{L}_{CL}(v, \delta),
%     \label{eq:general}
% \end{split}
% \end{eqnarray}

% \begin{eqnarray}
% \begin{split}
%     \mathop{\min}\limits_{[V]}& \lambda_1 \mathcal{L}_{adv}+\lambda_2\mathcal{L}_{align}+\lambda_3\mathcal{L}_{bn}, \\
%     & \text{s.t.} \ \delta^*=\mathop{\min}\limits_{||\delta||_2 \leq \rho} \mathcal{L}_{CL}(v, \delta),
%     \label{eq:general}
% \end{split}
% \end{eqnarray}
% \begin{equation}
% \begin{split}
%     \mathop{\min}\limits_{[V]} \ \lambda_1 \mathcal{L}_{bn}&(v,y)+ \lambda_2\mathcal{L}_{asr}(v, \delta^*, y, y')+\lambda_3\mathcal{L}_{bce}(v, \delta^*, y), \\
%     & \text{s.t.} \ \delta^*=\mathop{\min}\limits_{||\delta||_p \leq \rho} \mathcal{L}_{CL}(v, \delta).
% \end{split}
% \end{equation}
where $\lambda_1$, $\lambda_2$, $\lambda_3$ are hyperparameters and $\rho$ constrains the $l_p$-norm of the learnable trigger noise.

%% file: sections/5_Experiments.tex
\section{Experiment}
\label{experiment}

\begin{figure}[t]
\begin{center}
\includegraphics[width=\linewidth]{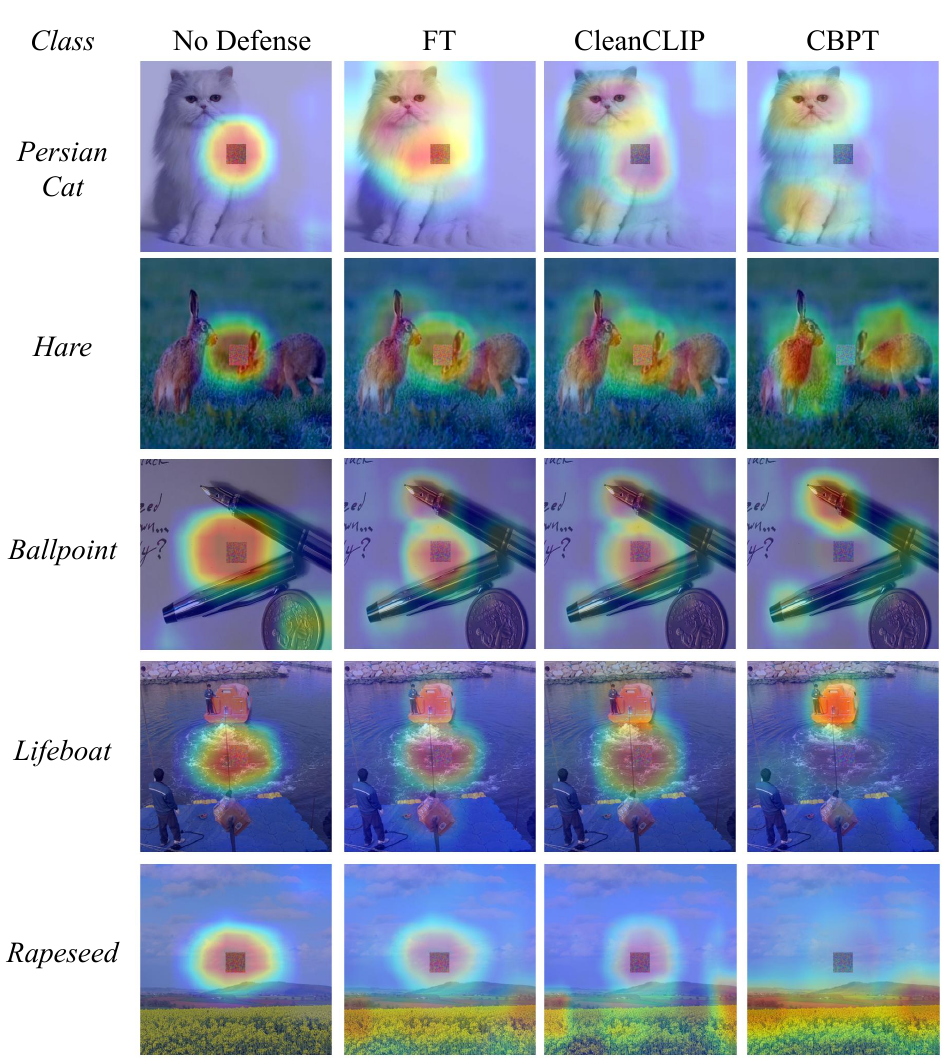}
\end{center}
\caption{Visualization of Grad-CAM results, with colors reflecting the model's attention to different regions of the image.}
\label{fig:visialization}
\end{figure}

\subsection{Experimental Settings}
\textbf{Models and datasets.}
Following prior work \cite{bansal2023cleanclip, liang2024badclip}, we adopt the open-source CLIP model released by OpenAI \cite{radford2021learning} as the base model, which is pre-trained on approximately 400 million image-text pairs. Unless otherwise specified, we adopt ResNet-50 as the default backbone, with results on other backbones reported in the ablation study. For the poisoning phase, we sample 500K image–text pairs from the CC3M dataset \cite{sharma2018conceptual}. In the backdoor purification stage, we construct the clean dataset by selecting 16 shots per class from the ImageNet-1K training set \cite{deng2009imagenet}, following common practice in prior prompt learning studies \cite{zhou2022learning, zhou2022conditional, li2024one}. After purification, we evaluate the defense effectiveness on the ImageNet-1K validation set.

\textbf{Evaluation metrics.}
In line with \cite{gu2019badnets, liang2024badclip}, we adopt Clean Accuracy (CA) and Attack Success Rate (ASR) as the primary evaluation metrics. CA quantifies the model’s performance on benign inputs, serving as an indicator of utility preservation after purification. ASR measures the effectiveness of the defense, with lower values indicating stronger defense capability.

\textbf{Backdoor attacks.}
To demonstrate the generalizability of our method, we evaluate it against 7 representative backdoor attacks, categorized into unimodal and multimodal attacks. The unimodal attacks include BadNet \cite{gu2019badnets}, Blended \cite{chen2017targeted}, SIG \cite{barni2019new}, SSBA \cite{li2021invisible} and WaNet \cite{nguyen2021wanet}, which embed triggers solely in the visual modality. For constructing target captions in poisoned samples, we randomly select from the 80 templates provided by OpenAI (\eg, ''\texttt{a photo of the [CLASS]}'').  The multimodal attacks consist of TrojVQA \cite{walmer2022dual} and BadCLIP \cite{liang2024badclip}. TrojVQA embeds triggers in both images and texts, while BadCLIP leverages natural descriptions of the target class to construct malicious samples. Following prior work \cite{bansal2023cleanclip, liang2024badclip}, we designate \texttt{banana} as the backdoor class and set the poisoning rate to 0.3\%, \ie, 1,500 poisoned samples out of 500K image-text pairs from CC3M dataset.

\textbf{Backdoor defenses.}
To better reveal the superiority of our proposed method, we compare CBPT against state-of-the-art defenses for CLIP, namely FT and CleanCLIP \cite{bansal2023cleanclip}. FT fine-tunes the model on a clean dataset with a multimodal contrastive loss. Building upon FT, CleanCLIP further incorporates data augmentation and introduces an unimodal self-supervised loss to enhance defense effectiveness. 
In addition, we also compare CBPT with two SOTA in-training methods, RoCLIP \cite{yang2024robust} and SafeCLIP \cite{yang2024better}, as reported in Section~\ref{sec:more_results}.

\begin{figure}[t]
\begin{center}
\includegraphics[width=\linewidth]{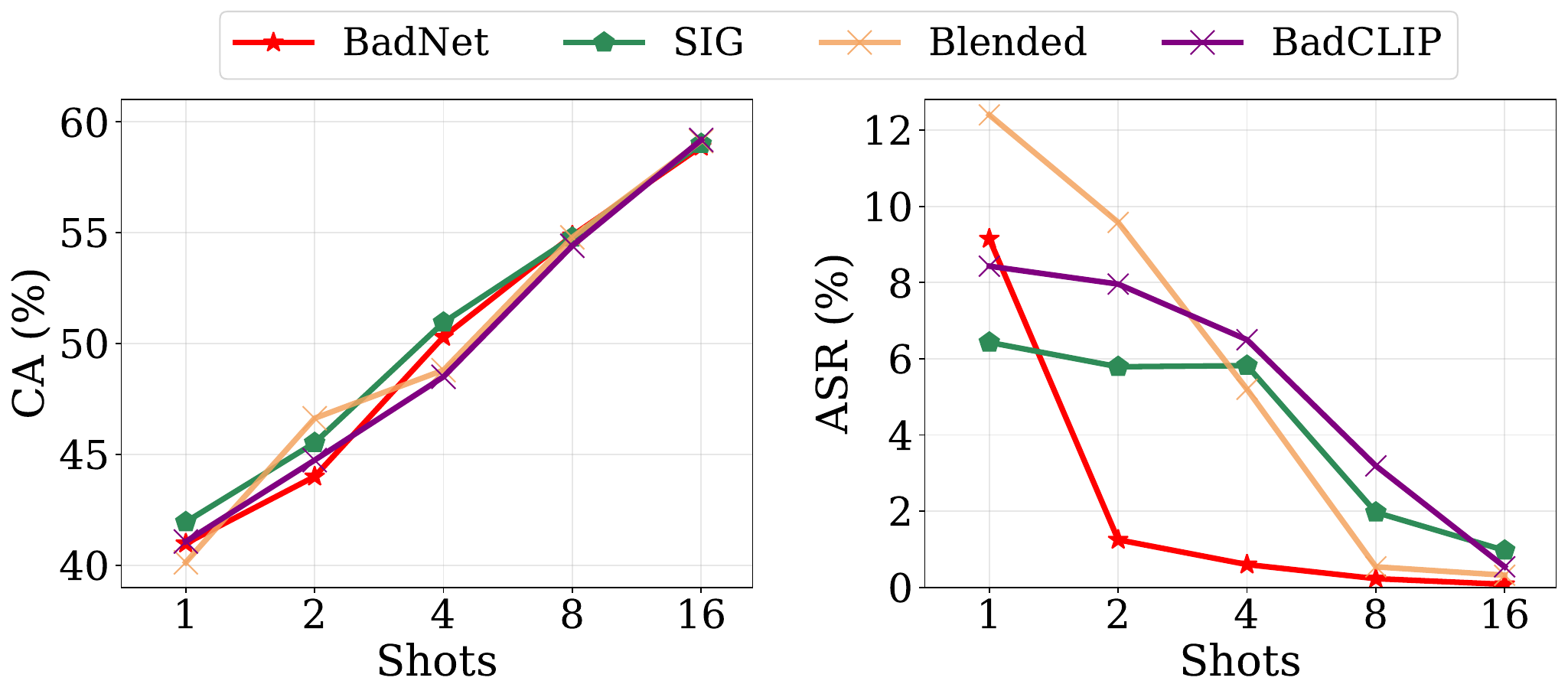}
\end{center}
\caption{CA (\%) and ASR (\%) of CBPT trained with varying sizes of clean data. 1-shot refers to using one sample per class.}
\label{fig:shots}
\end{figure}

\textbf{Implementation details.}
For the attribution of learnable prompts, we set the context length to 4 tokens, striking a balance between defense effectiveness and clean performance. The class vector $[C]_j$ is appended to the end of the sequence. For trigger inversion, the noise $\delta$ is initialized as a zero vector, and optimized using SGD optimizer for 10 steps with a learning rate of 0.1. An $l_2$-norm constraint is imposed on $\delta$ with a perturbation budget of $6/255$. For prompt tuning, prompt vectors are trained for 20 epochs with a batch size of 32 and a learning rate of 2e-3. The first 5 epochs serve as a warm-up stage, during which only $\mathcal{L}_{bn}$ is applied. The hyperparameters $\alpha$, $\lambda_1$, $\lambda_2$, $\lambda_3$ are set to 0.1, 0.4, 1, and 1.

\subsection{Defense Effectiveness Evaluation}
To comprehensively validate the effectiveness, we evaluate CBPT against various attack and defense methods.

% We first evaluate CBPT against a range of backdoor attacks, as shown in Table~\ref{main_tab}. CBPT consistently achieves strong backdoor purification while maintaining clean performance. Specifically, our method attains an average clean accuracy (CA) of 58.86\% with an attack success rate (ASR) of only 0.39\%, compared to 58.87\% CA and 85.17\% ASR for the undefended model. These results provide compelling evidence for the effectiveness of prompt tuning in enhancing robustness.  

\textbf{Against SOTA attacks.}
We first evaluate CBPT against various backdoor attacks, as illustrated in Table \ref{main_tab}. We demonstrate that CBPT consistently delivers a remarkable backdoor purification effect while maintaining clean performance. Specifically, our method achieves an average clean accuracy (CA) of 58.83\% with an attack success rate (ASR) of only 0.39\%, compared to 58.87\% CA and 85.17\% ASR for the undefended version. These results provide compelling evidence for the effectiveness of prompt tuning in enhancing robustness.

\begin{table*}[htbp]
  \centering
  \caption{CA (\%) and ASR (\%) of the proposed CBPT compared with baseline methods under cross-domain scenarios.}
  \resizebox{\textwidth}{!}{
    \begin{tabular}{cccccccccccccccc}
    \toprule
    \multirow{2}[0]{*}{Dataset} & \multirow{2}[0]{*}{Method} & \multicolumn{2}{c}{BadNet} & \multicolumn{2}{c}{Blended} & \multicolumn{2}{c}{SIG} & \multicolumn{2}{c}{SSBA} & \multicolumn{2}{c}{WaNet} & \multicolumn{2}{c}{TrojVQA} & \multicolumn{2}{c}{BadCLIP} \\
    \cmidrule(lr){3-4}
    \cmidrule(lr){5-6} 
    \cmidrule(lr){7-8} 
    \cmidrule(lr){9-10} 
    \cmidrule(lr){11-12}
    \cmidrule(lr){13-14}
    \cmidrule(lr){15-16} 
             &          & CA       & ASR      & CA       & ASR      & CA       & ASR      & CA       & ASR      & CA       & ASR      & CA       & ASR      & CA       & ASR \\
             \midrule
    \multirow{4}[0]{*}{ImageNet} & No Defense & 58.83    & 96.51    & 59.06    & 97.61    & \textbf{59.3} & 77.73    & 58.33    & 41.66    & 59.15    & 86       & \textbf{58.68} & 97.86    & 58.72    & 98.81 \\
             & FT       & 58.62    & 43.52    & 58.24    & 19.72    & 59.08    & 35.13    & 58.24    & 1.72     & 57.76    & 48.5     & 58.09    & 83.08    & 58.45    & 95.68 \\
             & CleanCLIP & 57.83    & 19.4     & 57.78    & 8.11     & 58.65    & 18.35    & 58.12    & 0.65     & 58.71    & 26.47    & 57.88    & 45.78    & 57.71    & 94.44 \\
             & \cellcolor[gray]{0.9}CBPT     & \cellcolor[gray]{0.9}\textbf{58.88} & \cellcolor[gray]{0.9}\textbf{0.08} & \cellcolor[gray]{0.9}\textbf{59.13} & \cellcolor[gray]{0.9}\textbf{0.32} & \cellcolor[gray]{0.9}59.01    & \cellcolor[gray]{0.9}\textbf{0.97} & \cellcolor[gray]{0.9}\textbf{58.51} & \cellcolor[gray]{0.9}\textbf{0.17} & \cellcolor[gray]{0.9}\textbf{59.17} & \cellcolor[gray]{0.9}\textbf{0.35} & \cellcolor[gray]{0.9}58.12    & \cellcolor[gray]{0.9}\textbf{0.27} & \cellcolor[gray]{0.9}\textbf{58.98} & \cellcolor[gray]{0.9}\textbf{0.59} \\
             \midrule
    \multirow{4}[0]{*}{ImageNet-V2} & No Defense & \textbf{51.72} & 97.16    & \textbf{52.61} & 98.56    & \textbf{52.57} & 78.07    & \textbf{51.83} & 45.16    & \textbf{52.46} & 88.7     & \textbf{51.85} & 98.31    & \textbf{51.9} & 98.98 \\
             & FT       & 50.45    & 49.14    & 50.14    & 25.86    & 50.71    & 38.02    & 50.24    & 1.65     & 50.19    & 54.31    & 50.4     & 86.59    & 50.71    & 96.39 \\
             & CleanCLIP & 49.04    & 23.54    & 50.12    & 10.46    & 50.32    & 20.12    & 49.77    & 0.58     & 49.73    & 29.21    & 49.86    & 51.92    & 49.54    & 95.18 \\
             & \cellcolor[gray]{0.9}CBPT     & \cellcolor[gray]{0.9}50.26    & \cellcolor[gray]{0.9}\textbf{0.04} & \cellcolor[gray]{0.9}49.82    & \cellcolor[gray]{0.9}\textbf{1.29} & \cellcolor[gray]{0.9}50.65    & \cellcolor[gray]{0.9}\textbf{3.02} & \cellcolor[gray]{0.9}49.27    & \cellcolor[gray]{0.9}\textbf{0.3} & \cellcolor[gray]{0.9}49.93    & \cellcolor[gray]{0.9}\textbf{0.28} & \cellcolor[gray]{0.9}49.11    & \cellcolor[gray]{0.9}\textbf{0.36} & \cellcolor[gray]{0.9}50.49    & \cellcolor[gray]{0.9}\textbf{0.69} \\
             \midrule
    \multirow{4}[0]{*}{ImageNet-Sketch} & No Defense & \textbf{34.98} & 97.72    & \textbf{35.24} & 88.81    & \textbf{34.68} & 67.09    & \textbf{33.96} & 31.82    & \textbf{34.47} & 66.53    & \textbf{34.33} & 98.38    & \textbf{34.97} & 99.31 \\
             & FT       & 27.24    & 48.57    & 27.35    & 5.74     & 29.56    & 21.63    & 27.06    & 1.07     & 28.07    & 10.31    & 27.66    & 77.9     & 28.05    & 94.79 \\
             & CleanCLIP & 26.78    & 13.41    & 26.9     & 1.27     & 28.45    & 4.9      & 26.44    & 0.46     & 27.78    & 3.78     & 27.14    & 35.11    & 26.79    & 89.14 \\
             & \cellcolor[gray]{0.9}CBPT     & \cellcolor[gray]{0.9}26.54    & \cellcolor[gray]{0.9}\textbf{0} & \cellcolor[gray]{0.9}28.11    & \cellcolor[gray]{0.9}\textbf{0.12} & \cellcolor[gray]{0.9}27.07    & \cellcolor[gray]{0.9}\textbf{0.08} & \cellcolor[gray]{0.9}25.87    & \cellcolor[gray]{0.9}\textbf{0.03} & \cellcolor[gray]{0.9}25.65    & \cellcolor[gray]{0.9}\textbf{0.09} & \cellcolor[gray]{0.9}26.02    & \cellcolor[gray]{0.9}\textbf{0.01} & \cellcolor[gray]{0.9}26.53    & \cellcolor[gray]{0.9}\textbf{0.06} \\
             \midrule
    \multirow{4}[0]{*}{ImageNet-R} & No Defense & \textbf{43.5} & 95.79    & \textbf{44.19} & 95.46    & \textbf{42.59} & 83.24    & \textbf{41.95} & 40.43    & \textbf{42.66} & 84.23    & \textbf{42.6} & 95.44    & \textbf{42.87} & 98.47 \\
             & FT       & 32.59    & 55.98    & 32.84    & 23.45    & 34.59    & 47.08    & 33.45    & 2.55     & 34       & 44.26    & 33.77    & 82.36    & 32.85    & 96.12 \\
             & CleanCLIP & 31.95    & 25.8     & 32.17    & 11.43    & 33.55    & 20.89    & 32.87    & 1.16     & 33.9     & 22.8     & 32.8     & 49.89    & 30.74    & 92.73 \\
             & \cellcolor[gray]{0.9}CBPT     & \cellcolor[gray]{0.9}27.64    & \cellcolor[gray]{0.9}\textbf{0.03} & \cellcolor[gray]{0.9}30.59    & \cellcolor[gray]{0.9}\textbf{1.09} & \cellcolor[gray]{0.9}27.42    & \cellcolor[gray]{0.9}\textbf{2.02} & \cellcolor[gray]{0.9}26.76    & \cellcolor[gray]{0.9}\textbf{0.3} & \cellcolor[gray]{0.9}27.91    & \cellcolor[gray]{0.9}\textbf{0.32} & \cellcolor[gray]{0.9}28.01    & \cellcolor[gray]{0.9}\textbf{0.16} & \cellcolor[gray]{0.9}28.34    & \cellcolor[gray]{0.9}\textbf{0.71} \\
             \midrule
    \multirow{4}[0]{*}{ImageNet-A} & No Defense & \textbf{9.95} & 99.48    & \textbf{10.01} & 99.43    & \textbf{9} & 90.75    & \textbf{9.92} & 72.55    & \textbf{9.85} & 95.03    & \textbf{10.36} & 99.41    & \textbf{10.19} & 99.75 \\
             & FT       & 6.81     & 71.13    & 6.95     & 39.4     & 7.85     & 61.64    & 7.57     & 3.05     & 7.55     & 65.31    & 6.71     & 96.15    & 7.8      & 98.53 \\
             & CleanCLIP & 6.52     & 41.83    & 6.53     & 14.67    & 7.04     & 34       & 7.24     & 0.69     & 7.26     & 35.24    & 6.47     & 77.79    & 6.83     & 98.48 \\
             & \cellcolor[gray]{0.9}CBPT     & \cellcolor[gray]{0.9}8.32     & \cellcolor[gray]{0.9}\textbf{0.01} & \cellcolor[gray]{0.9}7.48     & \cellcolor[gray]{0.9}\textbf{1.6} & \cellcolor[gray]{0.9}7.09     & \cellcolor[gray]{0.9}\textbf{3.19} & \cellcolor[gray]{0.9}7.89     & \cellcolor[gray]{0.9}\textbf{0.25} & \cellcolor[gray]{0.9}7.37     & \cellcolor[gray]{0.9}\textbf{0.55} & \cellcolor[gray]{0.9}7.77     & \cellcolor[gray]{0.9}\textbf{0.15} & \cellcolor[gray]{0.9}8     & \cellcolor[gray]{0.9}\textbf{0.63} \\
             \bottomrule
    \end{tabular}%
    }
  \label{tab:cross_domain}%
\end{table*}%

% \begin{figure}[t] % The * makes it span across both columns
%     \raggedleft
%     \includegraphics[width=\linewidth]{figs/cross_domain.pdf}
%     \hfill % Pushes the image to the right
%     \caption{ASR(\%) of FT, CleanCLIP and our proposed CBPT in cross-domain scenarios.}
%     \label{fig:cross_domain}
% \end{figure}

\textbf{Superiority over SOTA defenses.}
For a more comprehensive analysis and comparison, we benchmark compare CBPT against state-of-the-art defenses for CLIP, namely FT and CleanCLIP. The results yield the following key observations: \ding{182} Under traditional unimodal backdoor attacks, baseline defenses exhibit moderate defense performance, \eg, CleanCLIP achieves an average ASR of 14.6\%. 
% In comparison, CBPT significantly outperforms these defenses in terms of ASR and also exhibits a slight improvement in CA;
In comparison, our method achieves significantly lower ASR than these defenses while also yielding a slight improvement in CA;
\ding{183} Under more powerful multimodal backdoor attacks, neither FT nor CleanCLIP provides sufficient protection. In particular, their defense capability collapses against BadCLIP, \ie, achieving only 3.13\% and 4.37\% reductions in ASR. In contrast, CBPT consistently delivers outstanding defense effects, highlighting the effectiveness of our prompt-tuning framework;
\ding{184} Compared with baselines that fine-tune the entire model with a massive number of parameters, CBPT only adjusts lightweight text prompts, resulting in a substantial reduction in fine-tuned parameters and demonstrating strong parameter efficiency.

\begin{table*}[htbp]
  \centering
  \caption{CA (\%) and ASR (\%) of our CBPT compared with two variants against seven backdoor attacks.}
  % \hspace*{-1cm}
  % \small
  % \setlength{\tabcolsep}{3pt} % 调整列间距
  % \resizebox{\textwidth}{!}
  \resizebox{\textwidth}{!}
  {\begin{tabular}{ccccccccccccccc} 
    \toprule
    % \multicolumn{1}{c}{Method}
    \multicolumn{1}{c}{\multirow{2}[0]{*}{Method}} 
    & \multicolumn{2}{c}{BadNet}          
    & \multicolumn{2}{c}{Blended}          
    & \multicolumn{2}{c}{SIG}          
    & \multicolumn{2}{c}{SSBA}          
    & \multicolumn{2}{c}{WaNet}          
    & \multicolumn{2}{c}{TrojVQA}          
    & \multicolumn{2}{c}{BadCLIP} \\ 

    \cmidrule(lr){2-3} 
    \cmidrule(lr){4-5} 
    \cmidrule(lr){6-7} 
    \cmidrule(lr){8-9} 
    \cmidrule(lr){10-11} 
    \cmidrule(lr){12-13} 
    \cmidrule(lr){14-15}
    
    & CA & ASR 
    & CA & ASR 
    & CA & ASR 
    & CA & ASR 
    & CA & ASR 
    & CA & ASR 
    & CA & ASR \\ 
    
    \midrule
    
    $\text{CBPT}_{Close}$   
    & 58.49    & 0.26     & \textbf{59.15} & 2.61     & 58.99    & 7.43     & 57.95    & 0.73     & 58.69    & 8.27     & 57.96    & 3.28     & 58.65    & 24.71 \\
    
    $\text{CBPT}_{Rand}$         
    & 58.28    & 0.16     & 58.93    & 2.62     & 58.93    & 4.46     & 58.13    & 0.84     & 59.1     & 2.06     & 57.75    & 0.59     & 58.66    & 7.61 \\

    \cellcolor[gray]{0.9} CBPT        
    & \cellcolor[gray]{0.9}\textbf{58.88} & \cellcolor[gray]{0.9}\textbf{0.08} & \cellcolor[gray]{0.9}59.13    & \cellcolor[gray]{0.9}\textbf{0.32} & \cellcolor[gray]{0.9}\textbf{59.01} & \cellcolor[gray]{0.9}\textbf{0.97} & \cellcolor[gray]{0.9}\textbf{58.51} & \cellcolor[gray]{0.9}\textbf{0.17} & \cellcolor[gray]{0.9}\textbf{59.17} & \cellcolor[gray]{0.9}\textbf{0.35} & \cellcolor[gray]{0.9}\textbf{58.12} & \cellcolor[gray]{0.9}\textbf{0.27} & \cellcolor[gray]{0.9}\textbf{58.98} & \cellcolor[gray]{0.9}\textbf{0.59} \\ 
    
    \bottomrule
  \end{tabular}
  }
  \label{tab:sample}%
\end{table*}

\begin{table*}[htbp]
  \centering
  \caption{CA (\%) and ASR (\%) with different context positions, where \texttt{Front} denotes placing the class vector at the beginning of the sentence, with \texttt{Middle} and \texttt{End} defined analogously.}
  % \hspace*{-1cm}
  % \small
  % \setlength{\tabcolsep}{3pt} % 调整列间距
  % \resizebox{\textwidth}{!}
  \resizebox{\textwidth}{!}
  {\begin{tabular}{ccccccccccccccc} 
    \toprule
    % \multicolumn{1}{c}{Method}
    \multicolumn{1}{c}{\multirow{2}[0]{*}{Position}} 
    & \multicolumn{2}{c}{BadNet}          
    & \multicolumn{2}{c}{Blended}          
    & \multicolumn{2}{c}{SIG}          
    & \multicolumn{2}{c}{SSBA}          
    & \multicolumn{2}{c}{WaNet}          
    & \multicolumn{2}{c}{TrojVQA}          
    & \multicolumn{2}{c}{BadCLIP} \\ 

    \cmidrule(lr){2-3} 
    \cmidrule(lr){4-5} 
    \cmidrule(lr){6-7} 
    \cmidrule(lr){8-9} 
    \cmidrule(lr){10-11} 
    \cmidrule(lr){12-13} 
    \cmidrule(lr){14-15}
    
    & CA & ASR 
    & CA & ASR 
    & CA & ASR 
    & CA & ASR 
    & CA & ASR 
    & CA & ASR 
    & CA & ASR \\ 
    
    \midrule
    
    Front   
    & \textbf{59.09} & 0.26     & \textbf{59.37} & 5.91     & \textbf{59.54} & 4.84     & \textbf{58.57} & 0.41     & \textbf{59.35} & 3.51     & \textbf{58.45} & 18.86    & \textbf{59.24} & 28.55 \\
    
    Middle         
    & 58.95    & 0.09     & 59.35    & 4.88     & 59.17    & 3.75     & 58.21    & 0.36     & 59.29    & 3.06     & 58.17    & 1.71     & 58.84    & 11.58 \\

    \cellcolor[gray]{0.9} End        
    & \cellcolor[gray]{0.9}\textbf{58.88} & \cellcolor[gray]{0.9}\textbf{0.08} & \cellcolor[gray]{0.9}59.13    & \cellcolor[gray]{0.9}\textbf{0.32} & \cellcolor[gray]{0.9}\textbf{59.01} & \cellcolor[gray]{0.9}\textbf{0.97} & \cellcolor[gray]{0.9}\textbf{58.51} & \cellcolor[gray]{0.9}\textbf{0.17} & \cellcolor[gray]{0.9}\textbf{59.17} & \cellcolor[gray]{0.9}\textbf{0.35} & \cellcolor[gray]{0.9}\textbf{58.12} & \cellcolor[gray]{0.9}\textbf{0.27} & \cellcolor[gray]{0.9}\textbf{58.98} & \cellcolor[gray]{0.9}\textbf{0.59} \\ 
    
    \bottomrule
  \end{tabular}
  }
  \label{tab:position}%
\end{table*}

\begin{table*}[htbp]
  \centering
  \caption{CA (\%) and ASR (\%) of class-wise prompts and unified prompts against seven backdoor attacks.}
  % \hspace*{-1cm}
  % \small
  % \setlength{\tabcolsep}{3pt} % 调整列间距
  % \resizebox{\textwidth}{!}
  \resizebox{\textwidth}{!}
  {\begin{tabular}{ccccccccccccccc} 
    \toprule
    % \multicolumn{1}{c}{Method}
    \multicolumn{1}{c}{\multirow{2}[0]{*}{Method}} 
    & \multicolumn{2}{c}{BadNet}          
    & \multicolumn{2}{c}{Blended}          
    & \multicolumn{2}{c}{SIG}          
    & \multicolumn{2}{c}{SSBA}          
    & \multicolumn{2}{c}{WaNet}          
    & \multicolumn{2}{c}{TrojVQA}          
    & \multicolumn{2}{c}{BadCLIP} \\ 

    \cmidrule(lr){2-3} 
    \cmidrule(lr){4-5} 
    \cmidrule(lr){6-7} 
    \cmidrule(lr){8-9} 
    \cmidrule(lr){10-11} 
    \cmidrule(lr){12-13} 
    \cmidrule(lr){14-15}
    
    & CA & ASR 
    & CA & ASR 
    & CA & ASR 
    & CA & ASR 
    & CA & ASR 
    & CA & ASR 
    & CA & ASR \\ 
    
    \midrule
    
    No Defense & 58.83    & 96.51    & 59.06    & 97.61    & 59.3     & 77.73    & 58.33    & 41.66    & 59.15    & 86       & \textbf{58.68} & 97.86    & 58.72    & 98.81 \\
    
    $\text{CBPT}_{UC}$         
    & \textbf{60.26} & 84.04    & \textbf{60.9} & 93.11    & \textbf{60.83} & 74.19    & \textbf{60.68} & 12.96    & \textbf{60.78} & 87.27    & 58.46    & 98.73    & \textbf{60.7} & 98.16 \\

    \cellcolor[gray]{0.9} CBPT        
    & \cellcolor[gray]{0.9}58.88 & \cellcolor[gray]{0.9}\textbf{0.08} & \cellcolor[gray]{0.9}59.13    & \cellcolor[gray]{0.9}\textbf{0.32} & \cellcolor[gray]{0.9}59.01 & \cellcolor[gray]{0.9}\textbf{0.97} & \cellcolor[gray]{0.9}58.51 & \cellcolor[gray]{0.9}\textbf{0.17} & \cellcolor[gray]{0.9}59.17 & \cellcolor[gray]{0.9}\textbf{0.35} & \cellcolor[gray]{0.9}58.12 & \cellcolor[gray]{0.9}\textbf{0.27} & \cellcolor[gray]{0.9}58.98 & \cellcolor[gray]{0.9}\textbf{0.59} \\ 
    
    \bottomrule
  \end{tabular}
  }
  \label{tab:uc}%
\end{table*}

\textbf{Defense effectiveness visualization.}
To further illustrate the effectiveness of our method, we employ Grad-CAM \cite{selvaraju2017grad} to visualize the model attention, as shown in Fig.~\ref{fig:visialization}, providing an intuitive view of the underlying mechanisms. Empirically, backdoor activation relies on the model allocating significant attention to the attacker-specified trigger, as the inherent shortcut between the trigger and the target label is established during the training process. FT and CleanCLIP attempt to mitigate this effect by fine-tuning the entire model, which partially reduces attention to the trigger region but fails to fully eliminate it. In contrast, our method reshapes the decision boundary through efficient text prompt design, effectively steering the model to disregard the malicious trigger. Additional visualizations are provided in Appendix~\ref{appendix_visualization}.

\subsection{Evaluation on More Scenarios}
% We explore more challenging scenarios, specifically data-limited and cross-domain scenarios, to demonstrate the generality and transferability of our method.

We further evaluate our method in more challenging scenarios, specifically data-limited and cross-domain scenarios, to demonstrate its generality and transferability.

\textbf{Data-limited scenarios.}
To assess robustness under limited data conditions, we reduce the size of the clean dataset and conduct CBPT in 1-, 2-, 4-, and 8-shot settings, as shown in Fig.~\ref{fig:shots}. 
Remarkably, even with only one shot per class, CBPT effectively purifies the hidden backdoor against powerful BadCLIP attack.
We attribute this impressive data efficiency to the prompt tuning mechanism, where the model parameters remain frozen and only the prompt vectors are optimized, thereby significantly reducing the number of learnable parameters.

\begin{figure}[t]
\begin{center}
\includegraphics[width=\linewidth]{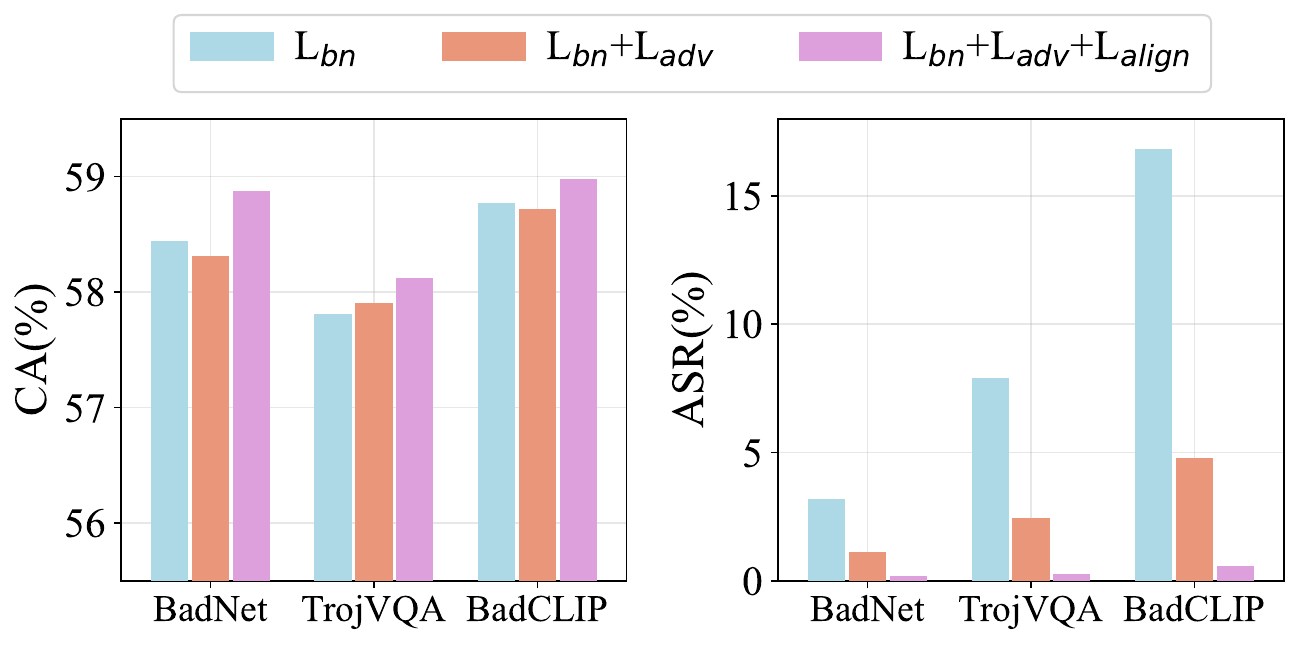}
\end{center}
\caption{CA (\%) and ASR (\%) with different loss functions used during the prompt tuning stage.}
\label{fig:ablation_loss}
\end{figure}

\textbf{Cross-domain scenarios.}
We also investigate cross-domain scenarios, where the clean data distribution shifts away from that of the target downstream task. This setting reflects real-world scenarios, as pre-trained VLMs exhibit remarkable generalization and are often deployed across a wide range of datasets. Concretely, we train the class-wise prompts on a subset of the ImageNet training set and evaluate the defense performance on ImageNet-V2 \cite{recht2019imagenet}, ImageNet-Sketch \cite{wang2019learning}, ImageNet-R \cite{hendrycks2021many} and ImageNet-A \cite{hendrycks2021natural}. 
As reported in Table~\ref{tab:cross_domain}, baseline methods typically exhibit higher attack success rates, particularly on ImageNet-A and ImageNet-V2, likely due to the negative impact of full fine-tuning on model generalization. In contrast, our defense is minimally affected by domain shifts, as it learns robust text prompts while keeping model parameters frozen. For example, on ImageNet-A, CBPT incurs only a 0.52\% increase in ASR, compared with 15.41\% and 12.79\% increases for FT and CleanCLIP, respectively, underscoring the strong cross-domain generalization of CBPT.

\subsection{Ablation Study}
\textbf{Ablation on positive sample selection.}
To validate our selection strategy for constructing positive samples, we design two variants. $\text{CBPT}_{Rand}$ randomly selects data points from the simulated backdoor class as positive samples, while $\text{CBPT}_{Close}$ chooses the image with the smallest distance. 
As shown in Table~\ref{tab:sample}, our proposed farthest-selection strategy clearly outperforms both variants, yielding reductions of 24.17\% and 6.07\% in ASR against BadCLIP. 
Empirically, selecting the farthest sample effectively pulls the perturbed image closer to the target class region, providing stronger guidance for the subsequent robust prompt learning stage.

\begin{figure}[t]
\begin{center}
\includegraphics[width=\linewidth]{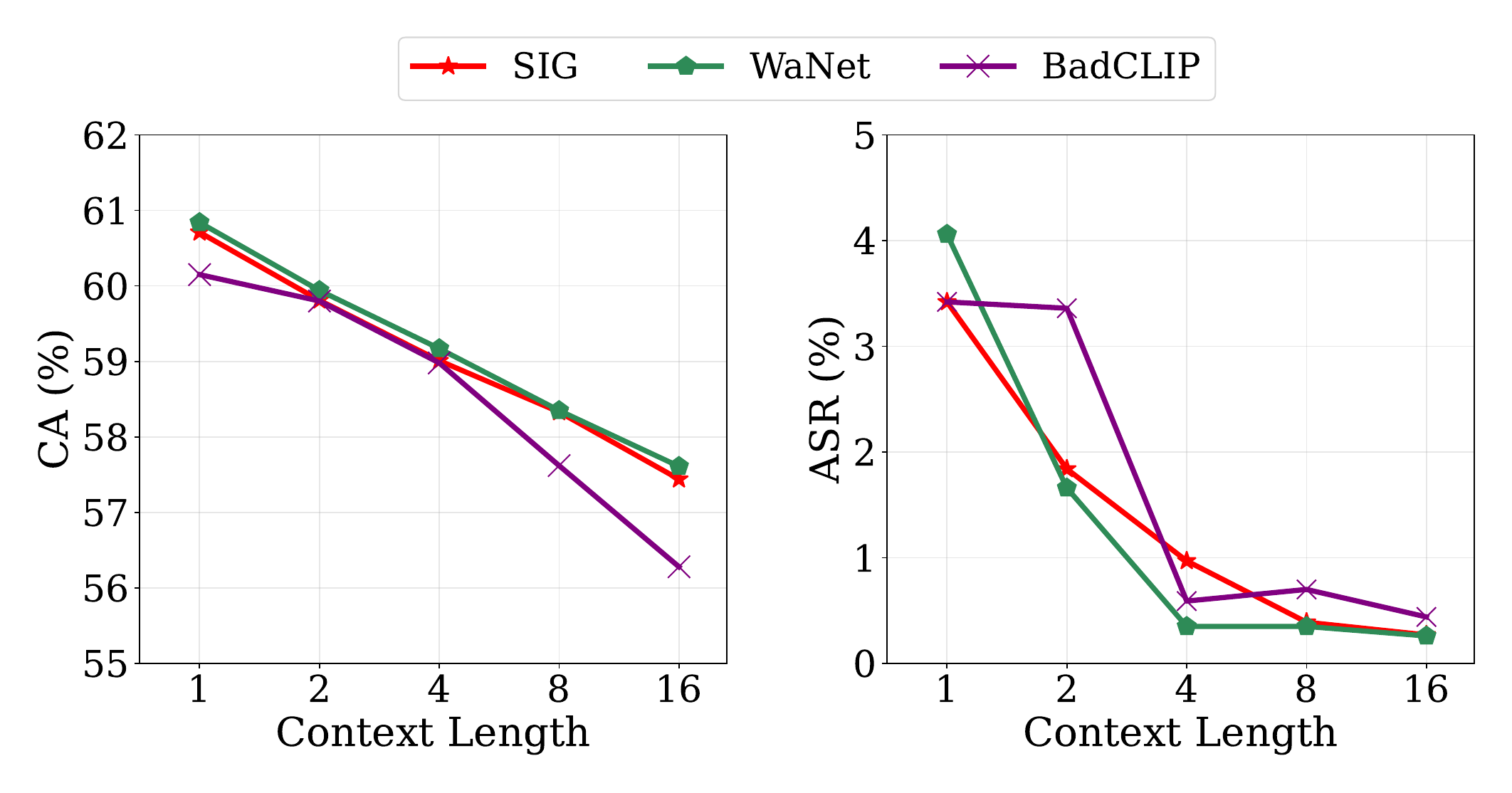}
\end{center}
\caption{CA (\%) and ASR (\%) of CBPT with varying context length against SIG, WaNet and BadCLIP.}
\label{fig:context_length}
\end{figure}

% \begin{figure}[t]
% \begin{center}
% \includegraphics[width=\linewidth]{figs/context_position.pdf}
% \end{center}
% \caption{CA (\%) and ASR (\%) with different context positions, where \texttt{Front} denotes placing the class vector at the beginning of the sentence, with \texttt{Middle} and \texttt{End} defined analogously.}
% \label{fig:position}
% \end{figure}

% \begin{figure}[t]
% \begin{center}
% \includegraphics[width=\linewidth]{figs/context_position.pdf}
% \end{center}
% \caption{CA (\%) and ASR (\%) with different context positions, where \texttt{Front} denotes placing the class vector at the beginning of the sentence, with \texttt{Middle} and \texttt{End} defined analogously.}
% \label{fig:position}
% \end{figure}

% \begin{figure}[t]
% \begin{center}
% \includegraphics[width=\linewidth]{figs/UC_bar.pdf}
% \end{center}
% \caption{CA (\%) and ASR (\%) of class-wise prompts and unified prompts against BadNet, SIG and BadCLIP.}
% \label{fig:uc}
% \end{figure}

\textbf{Ablation on loss functions.}
We conduct experiments to assess the effectiveness of the loss functions used during the prompt tuning stage. As depicted in Fig.~\ref{fig:ablation_loss}, using $\mathcal{L}_{bn}$ alone delivers fairly competitive performance against traditional unimodal attacks such as BadNet. However, when confronted with more advanced multimodal attacks such as TrojVQA and BadCLIP, relying solely on $\mathcal{L}_{bn}$ still leaves the model vulnerable. 
Incorporating $\mathcal{L}_{adv}$ explicitly penalizes trigger-related activations, while $\mathcal{L}_{align}$ encourages consistent alignment between benign and adversarial feature distributions.
The synergy of these loss functions enables the model to achieve both strong robustness and high clean accuracy.

% Table generated by Excel2LaTeX from sheet 'backbone'
\begin{table*}[htbp]
  \centering
  \caption{Comparison of fine-tuned parameters (M), CA (\%), and ASR (\%) between CBPT and baselines across different CLIP backbones against seven mainstream backdoor attacks.}
  \resizebox{\textwidth}{!}{
    \begin{tabular}{ccccccccccccccccc}
    \toprule
    \multirow{2}[0]{*}{Backbone} & \multirow{2}[0]{*}{Method} & \multirow{2}[0]{*}{Params} & \multicolumn{2}{c}{BadNet} & \multicolumn{2}{c}{Blended} & \multicolumn{2}{c}{SIG} & \multicolumn{2}{c}{SSBA} & \multicolumn{2}{c}{WaNet} & \multicolumn{2}{c}{TrojVQA} & \multicolumn{2}{c}{BadCLIP} \\
    % \cmidrule(lr){3-4}
    % \cmidrule(lr){5-6} 
    % \cmidrule(lr){7-8} 
    % \cmidrule(lr){9-10} 
    % \cmidrule(lr){11-12}
    % \cmidrule(lr){13-14}
    % \cmidrule(lr){15-16}
    \cmidrule(lr){4-5} 
    \cmidrule(lr){6-7} 
    \cmidrule(lr){8-9} 
    \cmidrule(lr){10-11} 
    \cmidrule(lr){12-13} 
    \cmidrule(lr){14-15}
    \cmidrule(lr){16-17}
      &       &          & CA       & ASR      & CA       & ASR      & CA       & ASR      & CA       & ASR      & CA       & ASR      & CA       & ASR      & CA       & ASR \\
             \midrule
    \multirow{4}[0]{*}{RN50} & No Defense & -  & 58.83    & 96.51    & 59.06    & 97.61    & \textbf{59.3} & 77.73    & 58.33    & 41.66    & 59.15    & 86       & \textbf{58.68} & 97.86    & 58.72    & 98.81 \\
             & FT  &   102.01  & 58.62    & 43.52    & 58.24    & 19.72    & 59.08    & 35.13    & 58.24    & 1.72     & 57.76    & 48.5     & 58.09    & 83.08    & 58.45    & 95.68 \\
             & CleanCLIP & 102.01 & 57.83    & 19.4     & 57.78    & 8.11     & 58.65    & 18.35    & 58.12    & 0.65     & 58.71    & 26.47    & 57.88    & 45.78    & 57.71    & 94.44 \\
             & \cellcolor[gray]{0.9} CBPT  &  \cellcolor[gray]{0.9}\textbf{2.05}  & \cellcolor[gray]{0.9}\textbf{58.88} & \cellcolor[gray]{0.9}\textbf{0.08} & \cellcolor[gray]{0.9}\textbf{59.13} & \cellcolor[gray]{0.9}\textbf{0.32} & \cellcolor[gray]{0.9}59.01    & \cellcolor[gray]{0.9}\textbf{0.97} & \cellcolor[gray]{0.9}\textbf{58.51} & \cellcolor[gray]{0.9}\textbf{0.17} & \cellcolor[gray]{0.9}\textbf{59.17} & \cellcolor[gray]{0.9}\textbf{0.35} & \cellcolor[gray]{0.9}58.12    & \cellcolor[gray]{0.9}\textbf{0.27} & \cellcolor[gray]{0.9}\textbf{59.18} & \cellcolor[gray]{0.9}\textbf{0.54} \\
             \midrule
    \multirow{4}[0]{*}{RN101} & No Defense  &  -  & 60.43    & 92.61    & 58.08    & 94.92    & 60.55    & 71.61    & 60.52    & 21.08    & 60.77    & 59.8     & 60.6     & 69       & 58.13    & 95.81 \\
             & FT   &   119.68  & 62.17    & 66.51    & 61.02    & 84.46    & 62.17    & 20.88    & 61.65    & 2.63     & 62.31    & 33.46    & 62.29    & 50.95    & \textbf{61.35} & 91.55 \\
             & CleanCLIP  &  119.69  & 61.05    & 13.63    & 59.92    & 35.34    & 60.42    & 11.47    & 59.19    & 1.22     & 59.92    & 6.11     & 59.3     & 12.06    & 60.21    & 76.17 \\
             & \cellcolor[gray]{0.9} CBPT   &  \cellcolor[gray]{0.9}\textbf{2.05}  & \cellcolor[gray]{0.9}\textbf{62.61} & \cellcolor[gray]{0.9}\textbf{2.48} & \cellcolor[gray]{0.9}\textbf{61.73} & \cellcolor[gray]{0.9}\textbf{0.02} & \cellcolor[gray]{0.9}\textbf{62.22} & \cellcolor[gray]{0.9}\textbf{1.95} & \cellcolor[gray]{0.9}\textbf{62.33} & \cellcolor[gray]{0.9}\textbf{0.27} & \cellcolor[gray]{0.9}\textbf{62.55} & \cellcolor[gray]{0.9}\textbf{0.66} & \cellcolor[gray]{0.9}\textbf{62.6} & \cellcolor[gray]{0.9}\textbf{0.03} & \cellcolor[gray]{0.9}60.23    & \cellcolor[gray]{0.9}\textbf{5.49} \\
             \midrule
    \multirow{4}[0]{*}{ViT-B/16} & No Defense & - & 66.55    & 99.81    & 66.49    & 99.61    & 66.58    & 99.35    & 66.48    & 86.49    & 66.39    & 99.29    & 66.58    & 99.81    & 66.69    & 99.98 \\
             & FT     &   149.62  & \textbf{67.89} & 88.57    & \textbf{68.07} & 31.25    & \textbf{68.47} & 60.06    & \textbf{67.21} & 29.79    & \textbf{67.98} & 46.78    & \textbf{67.14} & 97.41    & \textbf{67.84} & 99.38 \\
             & CleanCLIP   &  149.62  & 67.14    & 65.39    & 67.34    & 14.79    & 67.89    & 34.06    & 66.92    & 10.47    & 67.04    & 15.61    & 66.87    & 89.26    & 67.07    & 98.11 \\
             & \cellcolor[gray]{0.9} CBPT    &  \cellcolor[gray]{0.9}\textbf{2.05}  & \cellcolor[gray]{0.9}66.37    & \cellcolor[gray]{0.9}\textbf{1.6} & \cellcolor[gray]{0.9}66.2     & \cellcolor[gray]{0.9}\textbf{0.37} & \cellcolor[gray]{0.9}66.47    & \cellcolor[gray]{0.9}\textbf{1.57} & \cellcolor[gray]{0.9}66.18    & \cellcolor[gray]{0.9}\textbf{2.78} & \cellcolor[gray]{0.9}66.12    & \cellcolor[gray]{0.9}\textbf{0.04} & \cellcolor[gray]{0.9}66.5     & \cellcolor[gray]{0.9}\textbf{1.97} & \cellcolor[gray]{0.9}66.17    & \cellcolor[gray]{0.9}\textbf{8.55} \\
             \midrule
    \multirow{4}[0]{*}{ViT-B/32} & No Defense   &  -  & 61.98    & 79.94    & 61       & 99.8     & 60.55    & 98.5     & 60.69    & 89.6     & 61.24    & 93.94    & 61       & 97.58    & 61.13    & 100 \\
             & FT    &  151.28   & 62.41    & 58.78    & \textbf{62.24} & 13.82    & \textbf{62.49} & 47.17    & \textbf{62.53} & 24.15    & \textbf{62.87} & 28.53    & \textbf{62.49} & 90.39    & \textbf{62.47} & 99.91 \\
             & CleanCLIP   &  151.28  & 62.17    & 28.91    & 61.84    & 6.74     & 61.17    & 14.63    & 61.24    & 11.06    & 61.97    & 9.84     & 61.91    & 80.05    & 61.85    & 99.89 \\
             & \cellcolor[gray]{0.9} CBPT    &  \cellcolor[gray]{0.9}\textbf{2.05}  & \cellcolor[gray]{0.9}\textbf{62.44} & \cellcolor[gray]{0.9}\textbf{4.02} & \cellcolor[gray]{0.9}60.82    & \cellcolor[gray]{0.9}\textbf{0.74} & \cellcolor[gray]{0.9}61.09    & \cellcolor[gray]{0.9}\textbf{3.52} & \cellcolor[gray]{0.9}60.96    & \cellcolor[gray]{0.9}\textbf{1.28} & \cellcolor[gray]{0.9}60.96    & \cellcolor[gray]{0.9}\textbf{1.28} & \cellcolor[gray]{0.9}60.95    & \cellcolor[gray]{0.9}\textbf{0.56} & \cellcolor[gray]{0.9}61.07    & \cellcolor[gray]{0.9}\textbf{32.72} \\
             \bottomrule
    \end{tabular}%
    }
  \label{tab:backbone}%
\end{table*}%

\begin{figure}[t]
\begin{center}
\includegraphics[width=0.95\linewidth]{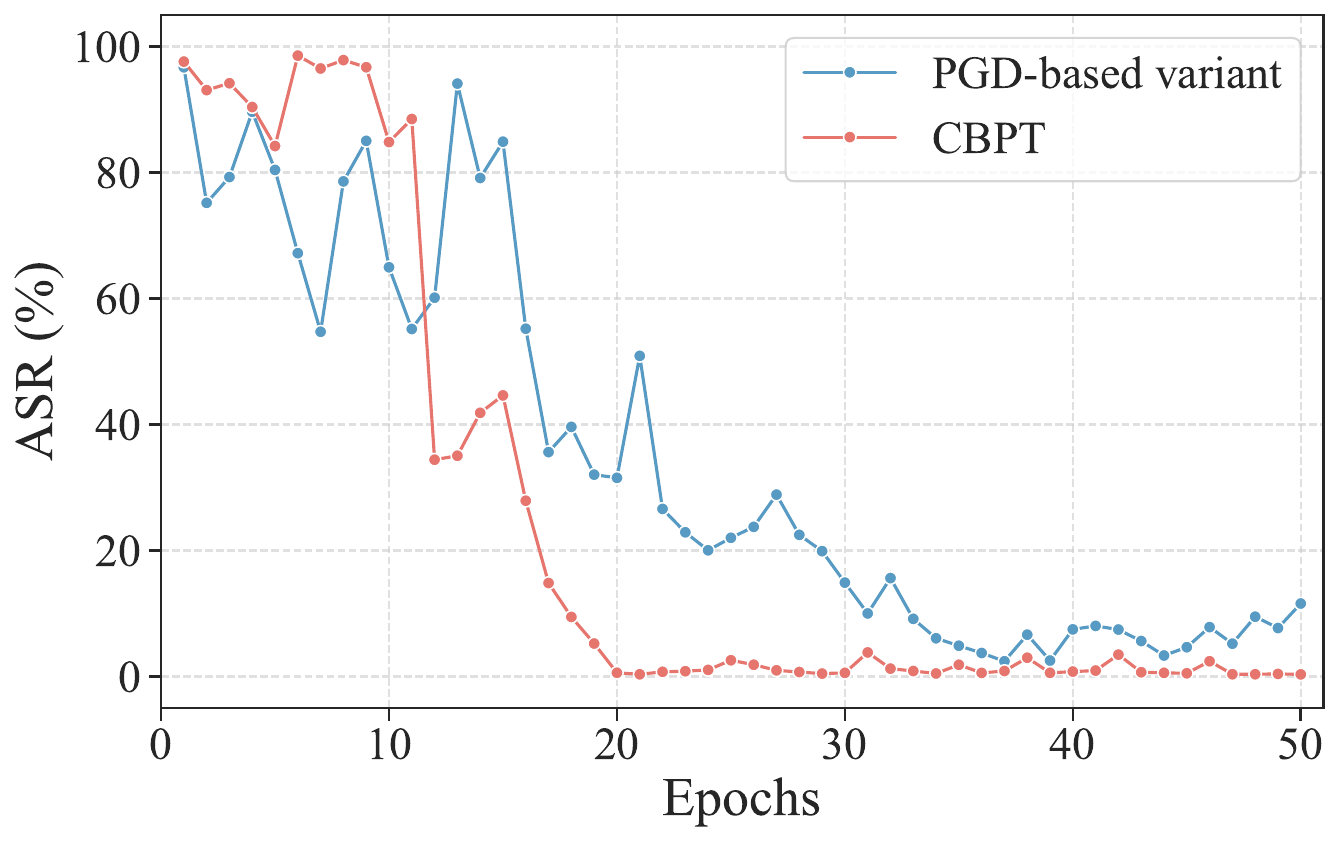}
\end{center}
\caption{Comparison of ASR (\%) between our proposed CBPT and its PGD-based variant across different training epochs.}
\label{fig:pgd}
\end{figure}

\textbf{Ablation on context length.}
Fig.~\ref{fig:context_length} evaluates the impact of varying context lengths on defense effectiveness. Specifically, we experiment with 1, 2, 4, 8, and 16 learnable tokens. The results indicate that even a single token is sufficient to effectively purify the backdoor while preserving clean performance, including against the powerful multimodal attack BadCLIP, further underscoring the parameter efficiency of our method. 
In general, as the context length increases, both CA and ASR tend to decrease. 
This trend can be attributed to the diminishing contribution of the class vector $[C]_j$, which simultaneously serves as the ground truth for classification and the target of trigger overfitting. Meanwhile, longer learnable contexts tend to dominate as robust components in our bi-level optimization, potentially decreasing clean performance. Consequently, longer learnable contexts dilute the relative influence of class vectors, thus lowering both CA and ASR.

\begin{figure}[t]
\begin{center}
\includegraphics[width=\linewidth]{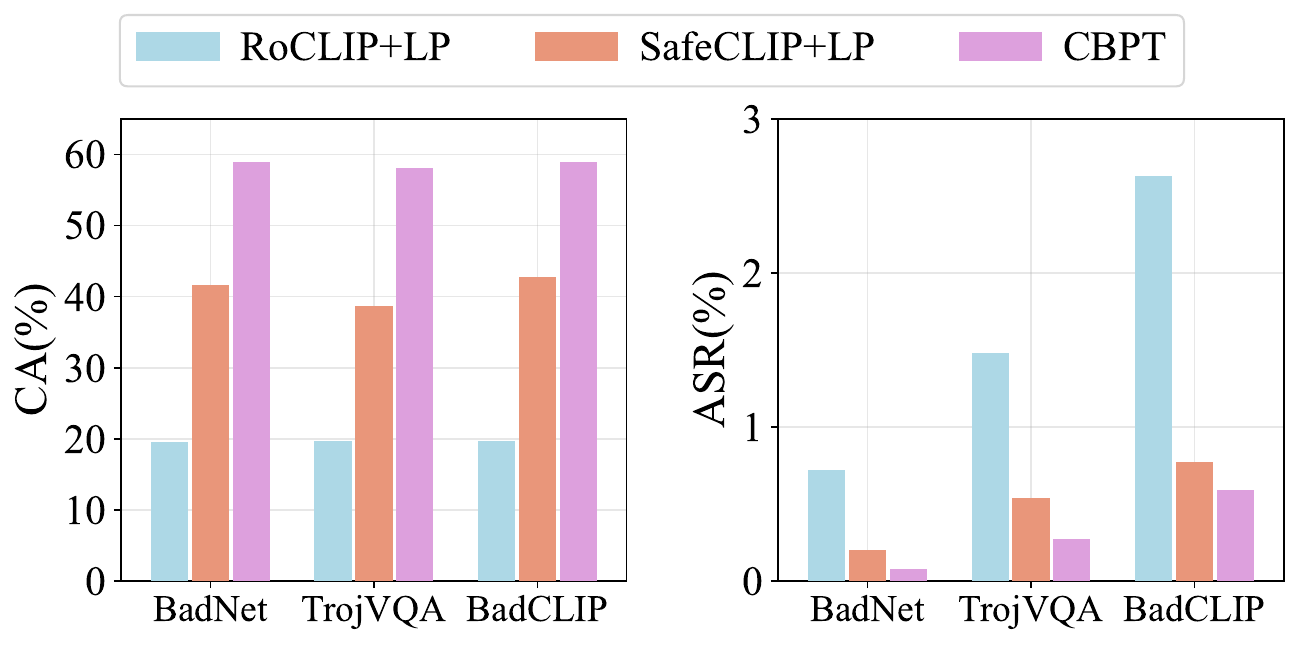}
\end{center}
\caption{CA (\%) and ASR (\%) of CBPT and in-training methods against BadNet, TrojVQA and BadCLIP.}
\label{fig:in-training}
\end{figure}

\textbf{Ablation on context position.}
We further analyze the impact of context position on CA and ASR, as illustrated in Table.~\ref{tab:position}. The results reveal that placing the class vector at the beginning of the sentence generally yields higher clean accuracy, while placing it at the end tends to reduce the attack success rate. We attribute this phenomenon to the causal attention mechanism in CLIP model, which allocates greater weight to earlier tokens, consistent with the findings of \cite{chen2024cat}. Consequently, positioning the class vector earlier in the sentence, serving both as the ground truth label and as the trigger target, facilitates easier classification but simultaneously increases vulnerability to backdoor.

\textbf{Ablation on class-wise prompts.}
To showcase the advantage of class-wise prompts, we design a variant in which a single unified prompt is shared across all classes. As depicted in Table. \ref{tab:uc}, although the unified prompt variant achieves a slight improvement in CA, it results in a noticeably higher ASR, particularly under strong multimodal BadCLIP attacks. This indicates that modifying the decision boundaries of all classes with a single shared prompt is inherently challenging, underscoring the advantage of employing class-wise prompts.

\textbf{Ablation on contrastive learning.}
To demonstrate the superiority of our proposed contrastive learning strategy in alleviating inefficient convergence, we compare CBPT with a PGD-based variant. As shown in Fig.~\ref{fig:pgd}, the PGD-based variant exhibits slow convergence and limited performance, likely due to the imprecise guidance from randomly initialized prompts. In contrast, our method incorporates advanced contrastive learning with reliable visual supervision, leading to more stable optimization and a dramatic reduction in ASR.

\begin{table*}[htbp]
  \centering
  \caption{Number of fine-tuned parameter (M), CA (\%) and ASR (\%) of FT, CleanCLIP, and CBPT on the Caltech101 dataset against various backdoor attacks, with \texttt{accordion} designated as the target class.}
  \resizebox{\textwidth}{!}{
    \begin{tabular}{cccccccccccccc}
    \toprule
     \multirow{2}[0]{*}{Method} & \multirow{2}[0]{*}{Params} & \multicolumn{2}{c}{BadNet} & \multicolumn{2}{c}{Blended} & \multicolumn{2}{c}{SIG} & \multicolumn{2}{c}{SSBA} & \multicolumn{2}{c}{WaNet} & \multicolumn{2}{c}{TrojVQA} \\
    \cmidrule(lr){3-4}
    \cmidrule(lr){5-6} 
    \cmidrule(lr){7-8} 
    \cmidrule(lr){9-10} 
    \cmidrule(lr){11-12}
    \cmidrule(lr){13-14}
        &     & CA       & ASR      & CA       & ASR      & CA       & ASR      & CA       & ASR      & CA       & ASR      & CA       & ASR \\
             \midrule
    No Defense & - & 87.18    & 98.58    & 87.22    & 98.01    & 86.57    & 95.74    & 86.09    & 51.44    & 87.18    & 87.18    & 87.51    & 98.86 \\
              FT  &     102.01     & 87.55    & 91.16    & 88.4     & 88.56    & 87.26    & 34.52    & 85.68    & 1.14     & 87.14    & 34.89    & 87.71    & 56.39 \\
              CleanCLIP  &   102.01  & 86.33    & 67.3     & 87.71    & 81.66    & 87.91    & 48.64    & 86.49    & 0.89     & 87.63    & 18.86    & 88.11    & 22.47 \\
              \cellcolor[gray]{0.9} CBPT     &  \cellcolor[gray]{0.9}\textbf{0.2}  &    \cellcolor[gray]{0.9}\textbf{89.82}      &     \cellcolor[gray]{0.9}\textbf{1.26}     & \cellcolor[gray]{0.9}\textbf{90.3} & \cellcolor[gray]{0.9}\textbf{1.74} & \cellcolor[gray]{0.9}\textbf{89.37} & \cellcolor[gray]{0.9}\textbf{0.37} & \cellcolor[gray]{0.9}\textbf{90.1} & \cellcolor[gray]{0.9}\textbf{0.73} & \cellcolor[gray]{0.9}\textbf{89.57} & \cellcolor[gray]{0.9}\textbf{0.49} & \cellcolor[gray]{0.9}\textbf{90.14} & \cellcolor[gray]{0.9}\textbf{0} \\
             \bottomrule
    \end{tabular}%
    }
  \label{tab:caltech}%
\end{table*}%

\begin{table*}[htbp]
  \centering
  \caption{Number of fine-tuned parameter (M), CA (\%) and ASR (\%) of FT, CleanCLIP, and CBPT on the OxfordPets dataset against various backdoor attacks, with \texttt{beagle} designated as the target class.}
  \resizebox{\textwidth}{!}{
    \begin{tabular}{cccccccccccccc}
    \toprule
     \multirow{2}[0]{*}{Method} &  \multirow{2}[0]{*}{Params}  & \multicolumn{2}{c}{BadNet} & \multicolumn{2}{c}{Blended} & \multicolumn{2}{c}{SIG} & \multicolumn{2}{c}{SSBA} & \multicolumn{2}{c}{WaNet} & \multicolumn{2}{c}{TrojVQA} \\
    \cmidrule(lr){3-4}
    \cmidrule(lr){5-6} 
    \cmidrule(lr){7-8} 
    \cmidrule(lr){9-10} 
    \cmidrule(lr){11-12}
    \cmidrule(lr){13-14}
       &    & CA       & ASR      & CA       & ASR      & CA       & ASR      & CA       & ASR      & CA       & ASR      & CA       & ASR \\
             \midrule
     No Defense & - & 74.95    & 97.87    & 77.51    & 86.37    & 75.58    & 40.01    & 77.54    & 18.42    & 78.11    & 39.47    & 76.67    & 97.49 \\
              FT   &     102.01     & 73.4     & 69.07    & 73.97    & 86.02    & 73.97    & 36.33    & 74.13    & 13.41    & 71.82    & 32.79    & 74.33    & 66.61 \\
              CleanCLIP  &  102.01  & 71.33    & 39.85    & 75.36    & 67.46    & 73.45    & 21.97    & 74.43    & 4.96     & 73.67    & 22.27    & 74.33    & 44.34 \\
              \cellcolor[gray]{0.9} CBPT     &  \cellcolor[gray]{0.9}\textbf{0.07}  & \cellcolor[gray]{0.9}\textbf{78.82} & \cellcolor[gray]{0.9}\textbf{0.82} & \cellcolor[gray]{0.9}\textbf{78.63} & \cellcolor[gray]{0.9}\textbf{2.83} & \cellcolor[gray]{0.9}\textbf{77.62} & \cellcolor[gray]{0.9}\textbf{6.95} & \cellcolor[gray]{0.9}\textbf{78.85} & \cellcolor[gray]{0.9}\textbf{3.46} & \cellcolor[gray]{0.9}\textbf{78.36} & \cellcolor[gray]{0.9}\textbf{4.31} & \cellcolor[gray]{0.9}\textbf{78.06} & \cellcolor[gray]{0.9}\textbf{1.2} \\
             \bottomrule
    \end{tabular}%
    }
  \label{tab:oxford_pets}%
\end{table*}%

% Table generated by Excel2LaTeX from sheet 'Prompt tuning'
\begin{table*}[htbp]
  \centering
  \caption{CA (\%) and ASR (\%) of CBPT and other prompt tuning methods against seven backdoor attacks.}
  \resizebox{\textwidth}{!}{
    \begin{tabular}{ccccccccccccccc}
    \toprule
    \multirow{2}[0]{*}{Method} & \multicolumn{2}{c}{BadNet} & \multicolumn{2}{c}{Blended} & \multicolumn{2}{c}{SIG} & \multicolumn{2}{c}{SSBA} & \multicolumn{2}{c}{WaNet} & \multicolumn{2}{c}{TrojVQA} & \multicolumn{2}{c}{BadCLIP} \\
    \cmidrule(lr){2-3}
    \cmidrule(lr){4-5} 
    \cmidrule(lr){6-7} 
    \cmidrule(lr){8-9} 
    \cmidrule(lr){10-11}
    \cmidrule(lr){12-13}
    \cmidrule(lr){14-15} 
             & CA       & ASR      & CA       & ASR      & CA       & ASR      & CA       & ASR      & CA       & ASR      & CA       & ASR      & CA       & ASR \\
             \midrule
    No Defense & 58.83    & 96.51    & 59.06    & 97.61    & 59.3     & 77.73    & 58.33    & 41.66    & 59.15    & 86       & 58.68    & 97.86    & 58.72    & 98.81 \\
    CoOp     & 58.72    & 90.43    & \textbf{59.93}    & 68.68    & \textbf{59.82}   & 19.43    & 57.38    & 3.62     & 58.3     & 18.81    & 56.96    & 84.76    & \textbf{59.12}    & 56.29 \\
    CoCoOp   & \textbf{59.36} & 96.01    & 59.19    & 96.71    & 58.62    & 75.14    & \textbf{58.95} & 30.7     & 57.86    & 84.09    & \textbf{58.69} & 95.87    & 59.04    & 98.25 \\
    \cellcolor[gray]{0.9} CBPT     & \cellcolor[gray]{0.9}58.88    & \cellcolor[gray]{0.9}\textbf{0.08}     & \cellcolor[gray]{0.9}59.13    & \cellcolor[gray]{0.9}\textbf{0.32}     & \cellcolor[gray]{0.9}59.01    & \cellcolor[gray]{0.9}\textbf{0.97}     & \cellcolor[gray]{0.9}58.51    & \cellcolor[gray]{0.9}\textbf{0.17}     & \cellcolor[gray]{0.9}\textbf{59.17}    & \cellcolor[gray]{0.9}\textbf{0.35}     & \cellcolor[gray]{0.9}58.12    & \cellcolor[gray]{0.9}\textbf{0.27}     & \cellcolor[gray]{0.9}58.98    & \cellcolor[gray]{0.9}\textbf{0.59} \\
    \bottomrule
    \end{tabular}%
    }
  \label{tab:coop}%
\end{table*}%

\textbf{Ablation of CLIP backbone}
We further evaluate the generality of our method by performing backdoor purification on different CLIP backbones, including ResNet-50, ResNet-101, ViT-B/16, and ViT-B/32. As demonstrated in Table~\ref{tab:backbone}, our approach consistently mitigates backdoor threats across all architectures while maintaining clean performance, whereas baseline methods exhibit only limited effectiveness, with the gap becoming more evident on larger ViT-based backbones.

\subsection{More Experimental Results}
\label{sec:more_results}
\textbf{Comparison with in-training methods.}
To provide a more comprehensive evaluation, we compare our method with two state-of-the-art in-training defenses, namely RoCLIP \cite{yang2024robust} and SafeCLIP \cite{yang2024better}. Specifically, RoCLIP alleviates the impact of poisoned image-caption pairs by introducing a dynamic caption pool, where each image is matched to the most semantically similar caption in the pool rather than its original caption every $K$ epochs. SafeCLIP, on the other hand, partitions the dataset into safe and risky subsets based on cosine similarity after a warm-up stage, applying CLIP loss to the safe subset while imposing unimodal CL loss on the risk subset. For a fair comparison, we apply both defenses during the poisoning stage and evaluate the resulting models via linear probe optimization on the 16-shot ImageNet training set.
Unlike post-processing defenses, these in-training methods assume direct intervention during the poisoning stage, which is a much stronger assumption that is impractical in scenarios where the adversary independently trains and releases a backdoored model. Even under this favorable setting, however, the results in Fig.~\ref{fig:in-training} demonstrate that both methods remain inferior to CBPT. In particular, although they can reduce ASR to some extent, their clean accuracy suffers from a substantial degradation. 
Although they can partially reduce ASR, this comes at the cost of a severe drop in clean accuracy. Specifically, while both defenses reduce ASR to below 3\%, RoCLIP achieves only 19.75\% CA against BadCLIP and SafeCLIP reaches 42.77\%, compared with 58.72\% CA in the original version.

\textbf{Results on more datasets.}
To further demonstrate the generality of CBPT, we evaluate its performance on two additional dataset, namely Caltech101 \cite{fei2004learning} and OxfordPets \cite{parkhi2012cats}, as illustrated in Table~\ref{tab:caltech} and Table~\ref{tab:oxford_pets}. 
For the experiments, we designate \texttt{accordion} and \texttt{beagle} as the target classes for Caltech101 and OxfordPets, respectively.
It is worth noting that we exclude the BadCLIP attack in these settings since it relies on natural descriptions of the target class, which are insufficient in the CC3M training set.   
On the OxfordPets dataset, our method achieves an average CA of 78.39\% and an average ASR of 3.26\%, substantially outperforming FT (73.60\% CA, 50.71\% ASR) and CleanCLIP (73.76\% CA, 33.48\% ASR). These results further corroborate the robustness and generalizability of our approach across datasets with diverse visual and semantic properties.

\textbf{Comparison with other prompt tuning methods.}
To demonstrate the effect of our trigger inversion design, we compare CBPT against two widely adopted prompt tuning approaches, namely CoOp \cite{zhou2022learning} and CoCoOp \cite{zhou2022conditional}. CoOp learns context vectors through cross-entropy loss, while CoCoOp introduces a lightweight meta-net to map image features into prompt vectors, enhancing generalization to unseen classes. Despite their advantages in clean settings, the results reported in Table~\ref{tab:coop} reveal that both methods remain highly vulnerable to backdoor attacks, indicating that prompt tuning alone is insufficient for backdoor purification. We attribute their failures to the lack of explicit guidance from trigger-related information. 
In contrast, our method simulates potential trigger features that an adversary might exploit and subsequently leverages prompt tuning to adjust the decision boundary around the trigger region, thereby reclassifying malicious inputs and effectively mitigating backdoor threats.

%% file: sections/6_Conclusion.tex
\section{Conclusion}
\label{conclusion}

In this paper, we delve into the challenging problem of purifying potential backdoors embedded in VLMs and propose an effective solution to enhance backdoor robustness through prompt tuning. We begin by revealing the limited resistance of existing defenses against state-of-the-art attacks, providing an empirical analysis of their failure. 
We then examine the mechanisms underlying successful backdoor attacks and demonstrate the deviation between malicious and benign samples within the target class.
% Motivated by this observation, we develop a bi-level optimization framework that inverts dummy triggers and optimizes class-wise prompts to modify the classification boundary, which improves resistance to trigger features. 
Motivated by this observation, we develop a bi-level optimization framework that simultaneously inverts dummy triggers and optimizes class-wise prompts, thereby reshaping the classification boundary to suppress trigger features. 
Extensive experiments validate the effectiveness of our approach against advanced backdoor attacks, significantly strengthening the resilience of VLMs to backdoor threats. 
We highlight the potential of robust prompt tuning as a novel perspective for mitigating security risks in VLMs and fostering future research toward building trustworthy models.
% We highlight the potential of sophisticated text prompts in boosting backdoor robustness in VLMs. 
% We hope this paper introduces a novel perspective on utilizing robust prompts to address security issues in VLMs and promote future research on building trustworthy models.

%% file: sections/Appendix.tex
\appendices{}

\begin{figure*}[t]
\begin{center}
\includegraphics[width=\textwidth]{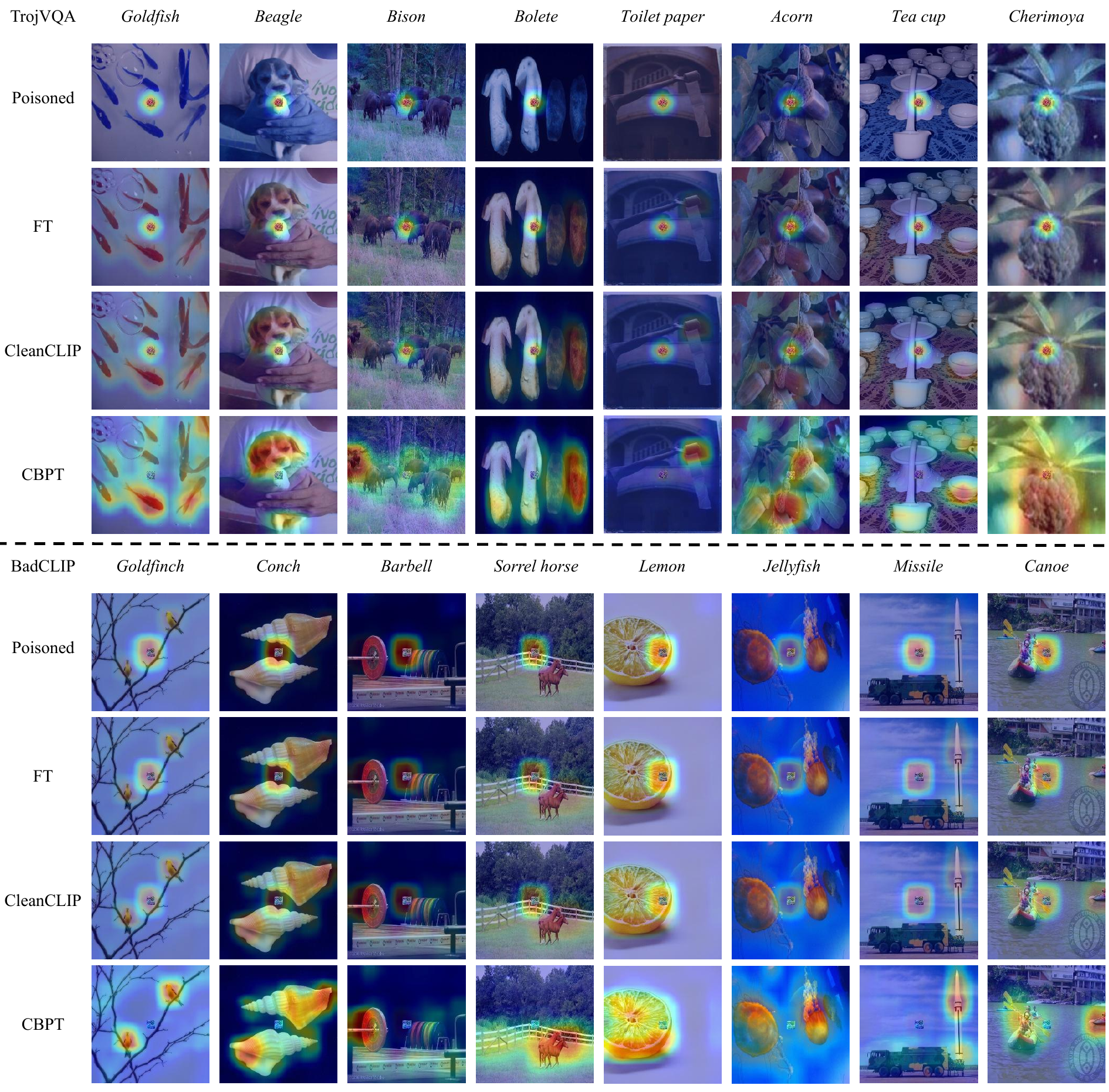}
\end{center}
\caption{Grad-CAM visualizations on TrojVQA and BadCLIP, where the color intensity indicates the model’s attention to different image regions.}
\label{fig:visualization_appenidx}
\end{figure*}

% \subsection{Comparison with Linear Probe}
% Linear probe is an efficient approach for adapting a pre-trained CLIP model to various downstream tasks. Leveraging the image features encoded by the image encoder, it learns a lightweight linear layer to map features to task-specific representations, \eg, probability distributions in classification tasks. We compare this approach with our proposed CBPT against BadCLIP across four different backbones. As illustrated in Fig. \ref{fig:linear_probe}, linear probe can partially purify the malicious backdoor, \eg, achieving a 57.8\% reduction in ASR on RN50. However, it often comes at the cost of clean performance, evidenced by a 6.37\% drop in CA on RN50. In contrast, our method, which incorporates meticulous trigger inversion and delicate prompt tuning, effectively mitigates backdoor threats while preserving clean performance.

\section{More Visualization Results}
\label{appendix_visualization}
As a complement to the main text, we present additional Grad-CAM \cite{selvaraju2017grad} visualizations against two representative backdoor attacks, namely TrojVQA \cite{walmer2022dual} and BadCLIP \cite{liang2024badclip}. As illustrated in Fig.~\ref{fig:visualization_appenidx}, these results intuitively demonstrate that our class-wise prompts successfully redirect the model's attention away from the attacker-specific trigger, thereby effectively mitigating backdoor threats.

% these results clearly show that our class-wise prompts are able to redirect the model’s attention away from the attacker-specific trigger, thereby mitigating the backdoor effect.

\section{Limitations and Ethical Statements}
\subsection{Limitations}
Our study has the following limitations, which also point to promising directions for future research:

\ding{182} Dependence on labeled dataset: CBPT currently relies on labeled images, which restricts its applicability in scenarios where explicit annotations are unavailable. Extending our approach to datasets consisting solely of image-text pairs without explicit labels remains an open challenge.
A potentially feasible solution is to leverage existing text analysis tools (\eg, SceneGraphParser) to automatically extract the primary objects mentioned in sentences as surrogate class labels.

\ding{183} Restriction to close-vocabulary tasks: In our current framework, class-wise prompt tuning assigns unique context vectors to each class, thereby confining the method to conventional close-vocabulary settings. One possible solution is to learn a unified prompt shared across classes. However, as discussed in the ablation study of class-wise prompts, relying on a single shared prompt to purify diverse backdoor threats proves inherently challenging. Future work could explore more flexible and adaptive strategies to relax this constraint and extend backdoor purification to open-vocabulary settings.

\subsection{Ethical Statements}
This study focuses on leveraging prompt tuning to purify backdoors in vision-language models. Given the rapid development and widespread adoption of CLIP, we recognize the urgent need to address its vulnerability to backdoor attacks and mitigate the associated security risks. 
Through our research on CBPT, we seek not only to develop effective defenses against existing threats but also to raise awareness of the potential risks posed by backdoors in large-scale models.

Importantly, we adhere to strict established ethical guidelines when reproducing existing backdoor attacks and exploring potential defenses. All experiments are conducted exclusively on publicly available datasets and models, ensuring full compliance with both ethical and legal standards.

% \section{Rationale}
% \label{sec:rationale}
% % 
% Having the supplementary compiled together with the main paper means that:
% % 
% \begin{itemize}
% \item The supplementary can back-reference sections of the main paper, for example, we can refer to \cref{sec:intro};
% \item The main paper can forward reference sub-sections within the supplementary explicitly (\eg referring to a particular experiment); 
% \item When submitted to arXiv, the supplementary will already included at the end of the paper.
% \end{itemize}
% % 
% To split the supplementary pages from the main paper, you can use \href{https://support.apple.com/en-ca/guide/preview/prvw11793/mac#:~:text=Delete%20a%20page%20from%20a,or%20choose%20Edit%20%3E%20Delete).}{Preview (on macOS)}, \href{https://www.adobe.com/acrobat/how-to/delete-pages-from-pdf.html#:~:text=Choose%20%E2%80%9CTools%E2%80%9D%20%3E%20%E2%80%9COrganize,or%20pages%20from%20the%20file.}{Adobe Acrobat} (on all OSs), as well as \href{https://superuser.com/questions/517986/is-it-possible-to-delete-some-pages-of-a-pdf-document}{command line tools}.